%% file: main.tex
\documentclass[dvipsnames, format=sigconf, anonymous=false, review=false, nonacm=true]{acmart} 
\AtBeginDocument{%
  \providecommand\BibTeX{{%
    \normalfont B\kern-0.5em{\scshape i\kern-0.25em b}\kern-0.8em\TeX}}}

\setcopyright{acmcopyright}
\copyrightyear{2018}
\acmYear{2018}
\acmDOI{10.1145/1122445.1122456}

\acmConference[Woodstock '18]{Woodstock '18: ACM Symposium on Neural
  Gaze Detection}{June 03--05, 2018}{Woodstock, NY}
\acmBooktitle{Woodstock '18: ACM Symposium on Neural Gaze Detection,
  June 03--05, 2018, Woodstock, NY}
\acmPrice{15.00}
\acmISBN{978-1-4503-XXXX-X/18/06}

\usepackage[ruled, lined, linesnumbered, commentsnumbered, noend]{algorithm2e}
\usepackage{xcolor}
\usepackage{soul}  
\usepackage{enumitem}

\usepackage{microtype}
\DeclareRobustCommand{\hlgray}[1]{{\sethlcolor{lightgray}\hl{#1}}}
\begin{document}

\input{notations}
\title{Understanding the Synergies between Quality-Diversity and Deep Reinforcement Learning}
\author{Bryan Lim}
\authornote{Both authors contributed equally to this research.}
\email{bryan.lim16@ic.ac.uk}
\affiliation{%
  \institution{Imperial College London}
  \city{London}
  \country{U.K.}
}

\author{Manon Flageat}
\authornotemark[1]
\email{manon.flageat18@ic.ac.uk}
\affiliation{%
  \institution{Imperial College London}
  \city{London}
  \country{U.K.}
}

\author{Antoine Cully}
\email{a.cully@imperial.ac.uk}
\affiliation{%
  \institution{Imperial College London}
  \city{London}
  \country{U.K.}
}

\renewcommand{\shortauthors}{Bryan Lim, Manon Flageat and Antoine Cully}

\begin{abstract}

The synergies between Quality-Diversity (QD) and Deep Reinforcement Learning (RL) have led to powerful hybrid QD-RL algorithms that have shown tremendous potential, and brings the best of both fields.
However, only a single deep RL algorithm (TD3) has been used in prior hybrid methods despite notable progress made by other RL algorithms.
Additionally, there are fundamental differences in the optimization procedures between QD and RL which would benefit from a more principled approach. 
We propose Generalized Actor-Critic QD-RL, a unified modular framework for actor-critic deep RL methods in the QD-RL setting.
This framework provides a path to study insights from Deep RL in the QD-RL setting, which is an important and efficient way to make progress in QD-RL.
We introduce two new algorithms, PGA-ME (SAC) and PGA-ME (DroQ) which apply recent advancements in Deep RL to the QD-RL setting, and solves the humanoid environment which was not possible using existing QD-RL algorithms.
However, we also find that not all insights from Deep RL can be effectively translated to QD-RL.
Critically, this work also demonstrates that the actor-critic models in QD-RL are generally insufficiently trained and performance gains can be achieved without any additional environment evaluations.

\end{abstract}



\begin{teaserfigure}
\centering
\vspace{-10pt}\includegraphics[width = 0.85\textwidth]{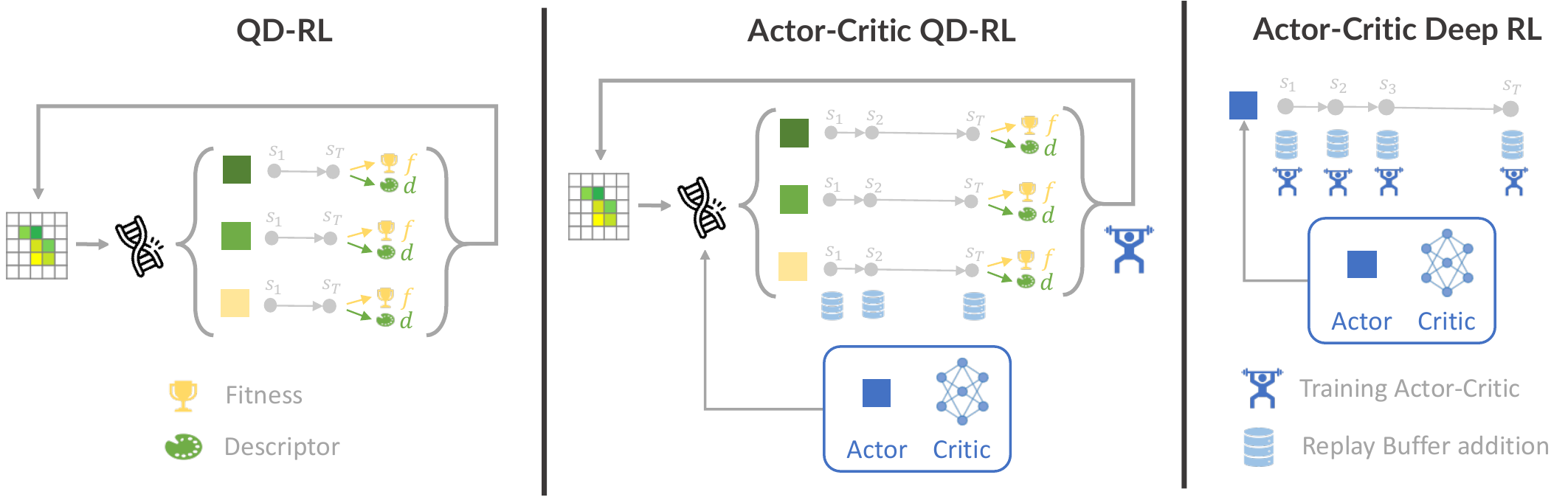}
\vspace{-8pt}\caption{
    Illustrative figure displaying the settings compared in this work: QD-RL, Actor-Critic QD-RL and Actor-Critic Deep RL.
    QD-RL refers generally to applying QD to RL domains to find collection of diverse policies, while Actor-Critic QD-RL approaches integrate elements from Actor-Critic Deep RL methods within QD to take advantage of the properties of the Deep RL domains. 
    This representation highlights key differences between Actor-Critic QD-RL and Actor-Critic Deep RL settings: 
    (1) update frequency of the actor-critic models and 
    (2) policies used to collect data.
}
\label{fig:featurefig}
\end{teaserfigure}

\maketitle

\section{Introduction}
Deep Reinforcement Learning (Deep RL) has made significant progress and demonstrated impressive breakthrough results across a wide range of application domains~\cite{silver2016mastering, akkaya2019solving, fawzi2022discovering, degrave2022magnetic}.
Similarly, Quality-Diversity (QD) algorithms have also shown to be effective, offering complementary characteristics such as diversity and goal switching which make them powerful optimization tools \cite{cully2015robots, ecoffet2021first, gaier2018data}.
QD algorithms can be used to find sets of diverse solutions to RL problems, which is commonly referred to as QD-RL.
While they can also be thought of as an alternative to Deep RL algorithms, combining QD and Deep RL synergistically has recently been explored and has demonstrated potential~\cite{ecoffet2021first, wang2019poet, nilsson2021policy, pierrot2022diversity}.
Prior work combined the TD3~\cite{fujimoto2018addressing} RL algorithm with QD (MAP-Elites) to provide gradient-informed variations in addition to the conventional mutation operators (PGA-ME~\cite{nilsson2021policy} and QD-PG~\cite{pierrot2022diversity}).
These have repetitively shown to be the best performing algorithms for the QD-RL setting~\cite{chalumeau2022neuroevolution,  tjanaka2022approximating}.


Given the promising nature of both these fields, it is critical to effectively translate the insights from Deep RL algorithms to QD-RL. 
Additionally, Deep RL is a large and fast moving field, evaluating its constant improvements is an efficient way to make progress in QD-RL.
However, existing QD-RL work has only ever used one type of Deep RL algorithm in the form of TD3 despite other algorithmic improvements in methods, such as SAC~\cite{haarnoja2018soft}, REDQ~\cite{chen2021randomized}, or DroQ~\cite{hiraoka2021dropout}.
This likely presents opportunities to further improve QD-RL algorithms if some of these insights and advancements can be translated.
Conversely, there are fundamental differences between the QD and Deep RL settings illustrated in Figure~\ref{fig:featurefig} that need to be considered and understood to realize the benefits of Deep RL in QD-RL.

\begin{figure}
\centering
\includegraphics[width = 0.45\textwidth]{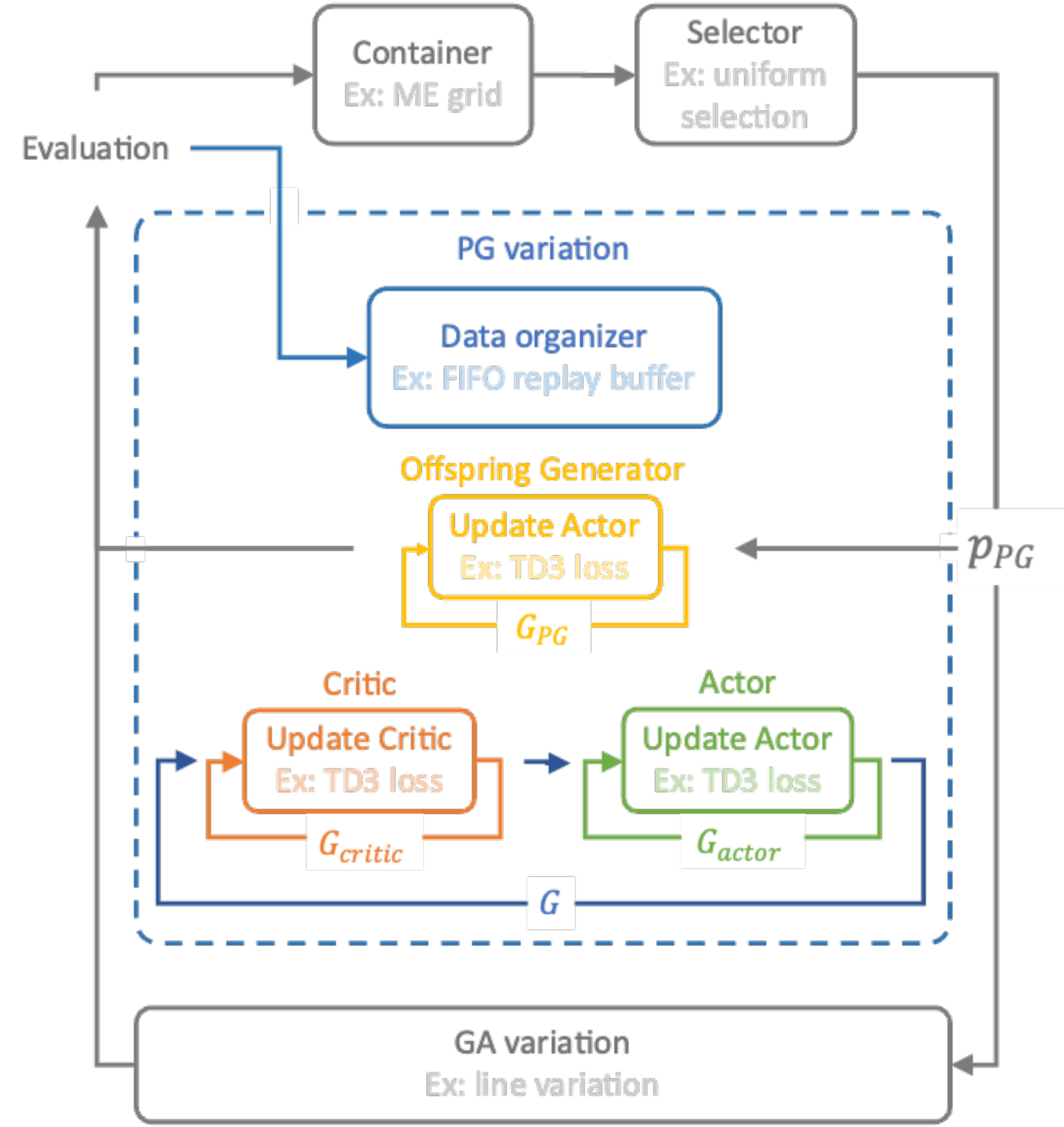}
\vspace{-8pt}\caption{
    Diagram of the \longname{} framework. Each solid box in the diagram is a module that can be easily swapped out to integrate latest advances in Deep RL.
    The main loop (gray) follows the conventional QD procedure~\cite{cully2017quality}. 
    Our generalized framework proposes an additional PG variation (blue). 
    Multiple modules constitutes the core of the framework: the Actor (green), Critic (orange), Offspring Generator (yellow) and Data Organizer to collect transitions from the environment (blue). 
}
\label{fig:gac}
\end{figure}

In this work, we propose \longname{}, a modular framework illustrated in Figure \ref{fig:gac}.
This framework aims to encourage advances in QD-RL algorithms by facilitating the evaluation of progress in Deep RL within QD-RL algorithms. 
Each module of this generalized framework can easily be swapped out to integrate new ideas and insights from Deep RL research and form new QD-RL algorithms. 
Additionally, \longname{} is also a powerful analysis tool that allows fundamental differences between the QD-RL and Deep RL settings to be studied.

Using \longname{}, we conduct an extensive study to better understand the synergies and differences between Deep RL and QD-RL. 
First, we introduce PGA-ME (SAC) and PGA-ME (DroQ) by switching up components of the Actor and Critic trainer modules in \longname{}.
Using the modularity of the framework, these algorithms separately evaluate three key advancements in Deep RL within QD-RL: (i) maximum entropy RL, (ii) increased critic training and (iii) critic regularization.
Through PGA-ME (SAC), we enable the Humanoid task to be solved for the first time in the QD-RL setting.
Interestingly, with PGA-ME (DroQ), we observe that not all advances in Deep RL can be translated easily to QD-RL. 
Notably, diversity of policies in QD-RL make regularization mechanisms from Deep RL redundant.
Second, the principled formalization of our framework leads us to reconsider the number of training steps of the actor and critic networks used in existing QD-RL algorithms. 
We find this number to be significantly under-estimated in prior QD-RL work and show that, by training these networks sufficiently, the performance of PGA-ME can be improved without any additional models or evaluations.
We demonstrate these results experimentally on three QD-RL tasks, using existing QD-RL algorithms as baselines.

\section{Related Work}

\subsection{Quality-Diversity}
Contrary to conventional optimization algorithms which search for a single objective-maximizing solution, Quality-Diversity (QD) algorithms search for a population of diverse and high-performing solutions~\cite{pugh2016quality, cully2017quality, chatzilygeroudis2021quality}.
The QD approach has found benefits in a wide range of applications such as robotics~\cite{cully2015robots}, video game design~\cite{gravina2019procedural}, engineering optimization~\cite{gaier2018data} and even discovery of drugs~\cite{verhellen2020illuminating} and materials~\cite{jiang2022chemical}.
QD algorithms are inspired by biological evolution where a diversity of behavioral niches are explicitly maintained during the optimization process. This process assigns to each solution $\theta$, a behavioral descriptor $d(\theta)$ value in addition to the task fitness or objective $F(\theta)$ value conventionally assigned to solutions in evolutionary computation. 
QD algorithms use local competition to only keep the best fitness solutions that fall within the same niche (i.e. have similar behavioral descriptors $d$). 
MAP-Elites (ME)~\cite{mouret2015illuminating}, a popular QD algorithm due to its simplicity, simplifies this process of competition by discretizing the behavioral descriptor space into an equally spaced grid.

QD optimization is commonly driven by gradient-free mutation operators as a result of its origins in evolutionary computation.
However, mutations are usually inefficient when dealing with larger search spaces, such as for applications in deep neuroevolution where the parameters of the neural network (weights and biases) are the search space.
To scale to higher-dimensional search spaces (thousands of parameters), a hybrid of gradient and evolutionary-based approaches have become more prominent.
ME-ES~\cite{colas2020scaling} explored this via natural gradients through the use of evolution strategies.
Another group of methods which are more sample efficient, consider the use of policy-gradient updates through the use of actor-critic methods from off-policy Deep RL algorithms~\cite{nilsson2021policy, pierrot2022diversity}.
Our work lies in the latter group of methods, more specifically building on the PGA-ME~\cite{nilsson2021policy} algorithm.

\subsection{Deep Reinforcement Learning}
Reinforcement Learning (RL) has long been studied as a way to solve sequential decision-making problems.
With the rise of deep learning, the powerful function approximation capabilities of neural networks have allowed RL methods to be successfully applied to a much wider range of problems. 
This is referred to as Deep RL.
The successes of Deep RL methods have then ranged over a wide number of applications and include achieving better than human-level performance on the Atari game suite~\cite{mnih2015human}, beating experts at complex games such as Go~\cite{silver2016mastering} and Starcraft~\cite{vinyals2019grandmaster}, robust legged locomotion~\cite{lee2020learning} and dexterous robotic manipulation~\cite{akkaya2019solving}.
More recently, Deep RL has also helped in making scientific breakthroughs in the fields of mathematics~\cite{fawzi2022discovering} and energy production~\cite{degrave2022magnetic}.

Despite its achievements, Deep RL still has some challenges that arise from the brittleness of learnt policies that demonstrate a lack of generalization. 
Ideas from QD and evolution have shown potential to aid in addressing these challenges.
For example, by bringing diversity to the behaviors learnt and tasks and environments used for learning.
Works inspired by QD have previously been used in Deep RL applications to achieve state-of-the-art performance on hard exploration tasks~\cite{ecoffet2021first} and also for pioneering work in automatic curriculum learning~\cite{wang2019poet, wang2020enhanced} for open-ended learning.

Our work utilizes the progress made in the field of Deep RL and studies how to effectively translate the improvements and progress made in this setting to the QD-RL setting.
We also look more specifically at certain key improvements made by the SAC~\cite{haarnoja2018soft} and DroQ~\cite{hiraoka2021dropout} algorithms which are off-policy Deep RL algorithms.
These algorithms are detailed in the next section.

\section{Preliminaries}

\subsection{Problem setting}

\paragraph{Reinforcement Learning}
The Reinforcement Learning (RL) problem is a setting in which an agent interacts with an environment and receives feedback in the form of a reward and a change in the state of the environment. The goal of the agent is to maximise this reward.
More formally, RL can be defined as a Markov Decision Process (MDP)~\citep{sutton2018reinforcement} consisting of the tuple $(\mathcal{S}, \mathcal{A}, \mathcal{P}, \mathcal{R})$, where $\mathcal{S}$ and $\mathcal{A}$ are the set of states and actions in the environment.
$\mathcal{P}(s_{t+1}|s_t, a_t)$ is the probability of transitioning from state $s_t$ to $s_{t+1}$ given the executed action $a_{t}$.
The reward function is used to represent the task we want the agent to learn and defines the reward obtained at each timestep $r_t=r(s_t, a_t, s_{t+1})$ when transitioning from state $s_t$ to $s_{t+1}$ under action $a_{t}$. 
A policy $\pi_{\theta}(a_{t} | s_t)$ determines how an agent selects its next action $a_{t}$ based on the current state $s_t$.
The objective in RL is then to optimize the parameters $\theta$ of policy $\pi_{\theta}$, such that it maximizes the expected cumulative reward $R(\tau)=\sum_{t=0}^T{r_t}$ over the entire episode trajectory $\tau = (s_0, a_0, ..., s_T, a_T)$ which corresponds to the states observed over the episode: 
$J(\pi_\theta) = \mathbb{E}_{\tau\sim\pi_\theta}\left[{\mathcal{R}(\tau)}\right]$.

\input{algorithms/deeprl_critic_training}

\paragraph{QD-RL}
We refer to QD algorithms used for RL problems as the QD-RL setting~\cite{nilsson2021policy, pierrot2022diversity, tjanaka2022approximating, flageat2022benchmarking}. In QD-RL, the fitness or objective value of a solution $F$ is defined as the total episodic reward and the descriptor $d$ is defined as a function of the state-action trajectory $\tau$:
\begin{align}
    F(\theta) = J(\pi_\theta) = \mathbb{E}_{\tau\sim\pi_\theta}\left[{\mathcal{R}(\tau)}\right] 
    \quad \text{and} \quad
    d(\theta) = \mathbb{E}_{\tau\sim\pi_\theta} \left[ d(\tau) \right]
\end{align}

In contrast to Deep RL which searches for a single objective maximising policy $\pi_\theta$, QD-RL searches for a population $\pop$ of diverse, objective maximising policies. PGA-ME~\cite{nilsson2021policy}, QD-PG~\cite{pierrot2022diversity} and CMA-MEGA \cite{tjanaka2022approximating} are examples of QD-RL algorithms which use Deep RL algorithms in the form of TD3~\cite{fujimoto2018addressing}.  

\subsection{Actor-Critic learning in Deep RL}
Most current Deep RL method relies on approximating the action-value function $Q(s_t, a_t)=\mathbb{E}_{\pi_{\theta}}\left[\sum_{k=0}^{T-t} \gamma^{k} r_{t+k+1} \mid s_t, a_t~\right]$, which encodes the expected return from following the policy $\pi_{\theta}$ after having performed action $a_t$ in state $s_t$. 
This approximation is done using a deep neural network $Q_{\psi}$ referred to as the critic network. 
The critic is used to improve the policy (or actor) $\pi_{\theta}$ by computing gradients.
Methods using both an actor network and a critic network are known as Actor-Critic.
Actor-Critic Deep RL approaches generally follow the procedure shown in Algorithm~\ref{alg:deeprl_critic_train}.
They commonly iterate between interacting with the environment using the actor policy $\pi_{\theta}$, learning the critic $Q_{\psi}$ for $G_{Q_{\psi}}$ gradient steps and then updating the actor policy $\pi_{\theta}$ for gradient $G_{\pi_{\theta}}$  steps.
Commonly used Actor-Critic Deep RL algorithms include notably TD3 \cite{fujimoto2018addressing} of which the update procedure for the critic and actor is detailed in Algorithm~\ref{alg:td3_training}. 
Previous work in QD-RL also almost exclusively employ TD3 as the Deep RL algorithm of choice~\cite{nilsson2021policy, pierrot2022diversity}. 
In this work, we study insights from alternative Actor-Critic Deep RL methods.

\input{algorithms/td3_update}

\section{Advancements in Deep RL}

This section provides overall background on the major Deep RL advancements that will be studied in the context of QD-RL.

\subsection{SAC: Maximum Entropy RL} \label{subsec:max-ent}
Overall, SAC is rather similar to TD3 as it is also model-free and off-policy. 
Soft-Actor Critic (SAC)~\cite{haarnoja2018soft} is an algorithm based on the maximum entropy RL framework. It introduces an additional entropy maximization term to the objective function on top of maximizing the reward (Equation~\ref{eqn:sac_obj}). The reason behind entropy maximization is to help the policy incentivise exploration and also possibly capture multiple modes of optimal behavior. 

\begin{align} \label{eqn:sac_obj}
    J(\pi_\theta) = \sum_{t=0}^T \mathbb{E}_{(s_t,a_t)\sim\pi_\theta}\left[{r(s_t, a_t) + \alpha\mathcal{H}(\pi(.|s_t)}\right]
\end{align}

SAC has a few key distinct differences to account for the new objective in Eq.~\ref{eqn:sac_obj}.
Firstly, SAC relies on stochastic policy for it to be able to perform the entropy maximization while TD3 uses deterministic policies. 
This means that each dimension of the action is modelled as a Gaussian distribution $N(\mu, \sigma)$, where the policy $\pi(|s)$ outputs parameters ($\mu$ and $\sigma$) of the distribution.
SAC also does not require any target policy smoothing, which involves artificially injecting noise to the actions' output by the policy in TD3.
This is because the maximum entropy objective and the stochastic policy provide the same regularization effect.
Policy and critic update procedures for SAC are provided in Algorithm~\ref{alg:sac_training}.

In terms of performance, SAC is known to be better performing and more reliable across a wider range of tasks. For example, SAC and TD3 both perform similarly across most of the benchmark continuous control gym tasks. However, SAC successfully enables a humanoid to walk while TD3 fails in this more complex environment. For this reason, SAC is more commonly used as a default benchmark and starting point for a lot of algorithms and applications in the RL community~\cite{chen2021randomized, hiraoka2021dropout, kumar2020conservative}.

\input{algorithms/sac_update}

\subsection{DroQ: Critic training and regularization} \label{subsec:critic_training}
While SAC is still commonly being used as the go-to state-of-the-art Deep RL baseline, there have also been improvements made on top of it to significantly increase sample efficiency.
These advancements have mainly been concerned with improving critic training.

The first insight that can be exploited is to increase the update-to-data ratio (UTD) when training the critic network~\cite{chen2021randomized}. 
The UTD ratio refers to how much the critic is trained when a new transition is collected in the environment (data). Thus, it corresponds to the number of critic gradient steps performed per new transition collected.
For clarity, the UTD only refers to the critic and not the actor. Hence, in the following, we refer to UTD as C-UTD (Critic-UTD).
More practically, when a new transition is collected, SAC usually trains both the actor and the critic a fixed number of steps (usually one step). 
Increasing the C-UTD ratio thus refer to increasing the ratio of number critic steps only while the number of actor steps remains the same. 
For a C-UTD of $20$, the critic will be trained $20$ times more than the actor when a new transition enter the replay buffer.
In algorithm~\ref{alg:deeprl_critic_train}, this refers to $G_{critic}$.

However, critically this increased C-UTD ratio has to be coupled with some form of regularization of the network to prevent overfitting of the critic given that more gradient steps will be taken.
REDQ~\cite{chen2021randomized} deals with this through a large ensemble of Q networks. However, this is known to be computationally expensive and takes large amounts of time.
DroQ~\cite{hiraoka2021dropout} addresses this computational cost by using Q-networks $Q_\psi$ regularized with dropout and layer normalization as a form of regularization to maintain the benefits of a high C-UTD ratio.
Smith et al.~\cite{smith2022walk} then showed that the general recipe of adding regularization to actor-critic RL methods is key to the jump in performance and a range of regularization methods can be used to obtain the same benefits.

\input{algorithms/gac.tex}

\section{Generalized Actor-Critic QD-RL} \label{sec:GAC_methods}

We present \longname{}, a modular framework, illustrated in Figure~\ref{fig:gac} and in Algorithm~\ref{alg:gac}.
Our framework propose a generalized view of actor-critic QD-RL methods to unify off-policy actor-critic Deep RL algorithms used in the QD-RL setting.
It builds on the QD framework~\cite{cully2017quality}, which defines a QD algorithm as the combination of Containers, Selectors and Variations.
Each component box in Figure~\ref{fig:gac} (\hlgray{highlighted lines} Algorithm~\ref{alg:gac}) is a module that can be easily swapped out. 
\longname{} introduces a Policy Gradient (PG) variation, which consists of the following modules:
\begin{itemize}[leftmargin=*]
    \item \textbf{Data Organizer:} store and manage collected transitions. 
    \item \textbf{Update Critic:} training procedure of the Critic, which can be changed depending on the Deep RL algorithm used.
    \item \textbf{Update Actor:} training procedure of the Actor which can be similarly changed depending on the Deep RL algorithm used.
\end{itemize}
On top, we also defined the following parameters, which are critical to encompass the full range of existing algorithms:
\begin{itemize}[leftmargin=*]
    \item \textbf{$G$:} collective number of critic and actor update loops at the end of the parallel episodic evaluations (at each generation).
    \item \textbf{$G_{critic}$:} specify the number of critic steps within one collective $G$ update (corresponds to $G_{critic}$ in Deep RLAlgorithm~\ref{alg:deeprl_critic_train}).
    \item \textbf{$G_{actor}$:} specify the number of actor steps within one collective $G$ update (corresponds to $G_{actor}$ in Deep RL in Algorithm~\ref{alg:deeprl_critic_train}).
    \item \textbf{$G_{PG}$:} specify the number of actor steps applied for each parent for PG offspring generation.
    \item \textbf{$p_{PG}$:} determines the proportion of policy-gradient variations relative to the entire batch size. This value is commonly divided equally between variations~\cite{nilsson2021policy, pierrot2022diversity}.
\end{itemize}

In the following sections, we first motivate the choice of these components, then we illustrate how the framework is used to create new algorithms. Finally, we show that this framework allows us to identify a limitation in current QD-RL training practices.

\subsection{Motivation for component modules} \label{subsec:motivation_methods}
We formulate \longname{} by principally considering the differences between the Deep RL and QD-RL setting (see Figure \ref{fig:featurefig}). We highlight some of these fundamental differences here and how the various components and parameters introduced in the framework aim to provide greater generalization.

\noindent \paragraph{\textbf{Off-policy actor-critic training.}}
The QD-RL setting evaluates a large number of different policies during optimization.
Thus, on-policy actor-critic methods such as TRPO or PPO~\cite{schulman2017proximal}, which can only update the policy on recent data collected by the current actor cannot be applied in the QD-RL setting.
The modules Update Actor and Update Critic, as well as parameters $G_{actor}$ and $G_{critic}$ encapsulate common off-policy actor-critic Deep RL algorithms. 
These components can be used to account for changes in loss functions and architectures.
They also allow the uniform implementation of features like the delayed policy updates (by changing $G_{critic}$ and $G_{actor}$) or high C-UTD settings of different Deep RL algorithms (by modifying $G_{critic}$), like DroQ.

\noindent \paragraph{\textbf{Timescales: Single-step vs Episodic}}
Most Deep RL algorithms usually function on the level of a time step also referred to as single transition. For example, in the single-policy Deep RL setting, algorithms SAC and TD3 update the critic and policy after one or a few environment time step(s) (see Algorithm \ref{alg:deeprl_critic_train} - line 9). 
On the other hand, population-based QD algorithms which have their origins in evolutionary computation consider episodic evaluations, only updates the critic and policies after an entire rollout of an episode (see Algorithm \ref{alg:gac} - line 21). 
The main reason behind this is that the behavioral descriptors and fitness values used in QD are defined as functions of the entire trajectory.
As a consequence, for an episode length of $T$, conventional Deep RL algorithms would have updated the critic and policy at least $T$ times. 
On the contrary, QD-RL only update the networks at the end of the parallel evaluations of policies, leading to a discrepancy in total number of training steps taken.

Parameter $G$ introduced in our framework aims to compensate for this by controlling the number of update loops after each generation. This allows the values of $G_{critic}$ and $G_{actor}$ to be maintained and follow the same training procedure the Deep RL algorithm of choice uses per timestep.

\noindent \paragraph{\textbf{Data Distribution: Single-policy vs Population of policies}}
As highlighted in Figure~\ref{fig:featurefig}, in QD-RL, data comes from $B$ different policies, which all exhibit different behaviors. 
On the contrary, in most Deep RL approaches, transitions are collected by a single actor. 
As a consequence, there is a significant difference in distribution of the replay buffer $\mathcal{D}$ data in QD-RL compared to Deep RL that needs to be considered.
This issue is further amplified by the previous point highlighting that the actor is usually updated between transitions in Deep RL.
This results in a biased data distribution towards the reward maximising actor policy.
In contrast, the batch $B$ policies evaluated in QD-RL stay constant for the length of the episode, leading to even more severe distribution changes in $\mathcal{D}$.
Additionally, the data contained in the replay buffer $\mathcal{D}$ will be swapped out at a higher frequency in QD-RL than in Deep RL, and be fully replaced after just a few generations.

The Data Organizer module can account for this through the type (e.g., fifo) and size of replay buffer used, and sorting the source of the transitions respectively. It also allows the flexibility of various sampling procedures in Deep RL like importance sampling.

\subsection{Using Generalized Actor-Critic QD-RL to integrate advances in Deep RL}

Using the generalized Generalized Actor-Critic framework, we take two established and proven Deep RL algorithms and improvements different to TD3 (detailed in Section~\ref{subsec:max-ent} and~\ref{subsec:critic_training}), in the form of SAC and DroQ and propose variants of PGA-ME which we refer to as PGA-ME (SAC) and PGA-ME (DroQ).

\subsubsection{PGA-ME (TD3)} \label{sec:pga_td3} The original PGA algorithm can be recovered by using the TD3 procedures (in Algorithm~\ref{alg:td3_training}) for the Update Critic and the Update Actor modules.
$G_{critic}$ is set to $2$ and $G_{actor}$ to $1$ to implement the delayed policy updates, while $G$ is set to $150$ to recover $300$ critic steps as in author implementation~\cite{nilsson2021policy}.

\subsubsection{PGA-ME (SAC)} \label{sec:pga_sac} PGA-ME (SAC) is implemented by replacing the Update Critic and Update Actor from the TD3 procedure with the SAC update procedure in Algorithm~\ref{alg:sac_training}. 
This includes a new loss function and max. entropy objective, while also using stochastic policies (see Section~\ref{subsec:max-ent}).

\subsubsection{PGA-ME (DroQ)} \label{sec:pga_droq} As DroQ builds on the SAC algorithm, most of the same building blocks from \PGASAC{} can be used. 
In PGA-ME (DroQ), the high C-UTD setting is implemented by taking $G_{actor}=1$ and $G_{critic} = C-UTD > 1$ to train the critic more at each generation.
Additionally, the Critic Update module is modified to use dropout and layer normalization.

\subsection{Consequences for QD-RL approaches} \label{subsec:methods_study}

In this section, we highlight the benefit of our Generalized Actor-Critic framework to study current practices in QD-RL, using the principled considerations of the differences between Deep RL and QD-RL settings (Section~\ref{subsec:motivation_methods}).

For an Actor-Critic QD-RL algorithm, $T \times B$ transitions would have been collected in the environment by $B$ distinctly different policies and added to the replay buffer between each update of the critic $Q_\psi$ and actor $\pi_{\theta_a}$ (after each generation).
In the Generalized Actor-Critic framework, the parameter $G$ corresponds to the total number of update loops of $Q_\psi$ and $\pi_{\theta_a}$ and allows (1) transitions in the replay buffer to be sampled more, and (2) critic and actor to be trained more (i.e. more gradient steps)

A common value for $T$ would be $1,000$~\cite{tassa2018deepmind, brax2021github} while $B$ would be $128$~\cite{cully2015robots, nilsson2021policy, tjanaka2022approximating} translating into on the order of $\sim 100,000$ of transitions before updating the critic and actor.
In common Deep RL settings, this would require $G \sim 50,000$ collective critic and actor update loops.
In comparison, the author's implementation of PGA-ME chose hyper-parameter values that corresponds to $G=150$, which is several orders of magnitude smaller. 
This suggests that previous work in QD-RL largely underestimate the value of the total number of updates $G$.
We study the effect of this difference by increasing $G$ in the following experiments. 
This study is distinct from studying high C-UTD in PGA-ME (DroQ), as C-UTD corresponds to the number of critic updates per collected transition, so is controlled by $G_{critic}$; while here we are proposing to study the effect of the collective number of critic and actor steps, $G$.

\section{Experiments}

To better understand the relationship between QD and Deep RL algorithms, we experimentally study the methods and insights enabled by our Generalized Actor-Critic framework. 
Following our analysis in previous sections, we aim to answer the following:
\begin{enumerate}[leftmargin=*]
    \item Can the advances in Deep RL be translated to QD-RL, namely:
    \begin{itemize}[leftmargin=*]
        \item maximum entropy RL through PGA-ME (SAC) (Section~\ref{sec:pga_sac})?
        \item C-UTD through PGA-ME (DroQ - Reg. + C-UTD) (Section~\ref{sec:pga_droq})?
        \item critic regularization through PGA-ME (DroQ - Reg.) (Section~\ref{sec:pga_droq})?
    \end{itemize}
    \item Is the intuition brought by our formalization, that actor-critic networks are usually under-trained in QD-RL correct?
\end{enumerate}
These studies highlight the importance of our framework for applying future advancements in Deep RL algorithms to QD-RL. 

\subsection{Experimental Setup}
\begin{figure}[t!]
\centering
\includegraphics[width = 0.4\textwidth]{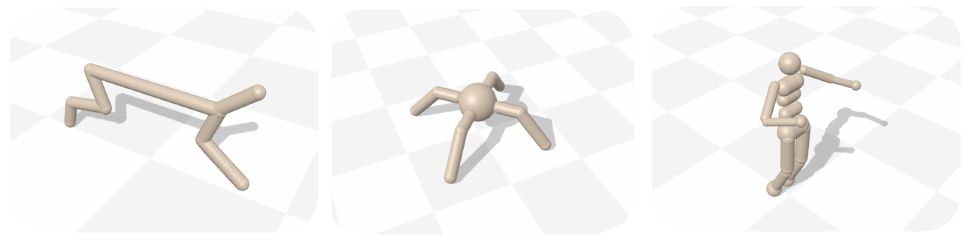}
\vspace{-10pt}
\caption{
    The Halfcheetah, Ant and Humanoid environments.
}
\label{fig:envs}
\end{figure}

For our study, we use continuous control QD-RL benchmark tasks~\cite{flageat2022benchmarking} in the form of the Half-cheetah, Ant and Humanoid on the Brax~\cite{brax2021github} simulator (see Figure~\ref{fig:envs}). These tasks were also used in the original PGA-ME paper~\cite{nilsson2021policy}. However, we also add the Humanoid task to evaluate our proposed methods.
All these tasks are uni-directional, where the goal is to find a diverse set of gait policies $\pop$ to move forward as fast as possible.
This definition is inspired and shown to be effective for applications in rapid adaptation and damage recovery~\cite{cully2015robots}.
For a policy $\theta \in \pop$, the fitness $F(\theta)$ and descriptor $d(\theta)$ are defined as follow:

\vspace{-11pt}
\begin{align}
    d(\theta) &= \frac{1}{T} \sum_{t}^{T}{
    \begin{pmatrix}
    C_1^\theta(t) \\
    \vdots \\
    C_I^\theta(t) 
    \end{pmatrix}\textrm{, where $I$ is the number of feet.}
    } \\
    F(\theta) &= \sum_{t=0}^{T}{r_{forward}(\theta) + r_{survive}(\theta) + (-r_{torque}(\theta))}
\end{align}

where $C_i^\theta$ refers to a boolean which is 1 when the foot $i$ is in contact with the ground and 0 when not. 
This choice of behavioral descriptor gives rise to diverse gaits as it quantifies the average usage of each leg when walking.
All environments have an episode length $T=1000$.
For all PGA-ME variants considered, we use a batch size $B$ of $128$ and we perform $5$ replications of each run. We compute p-values based on the Wilcoxon rank-sum test with a Bonferroni correction for all metrics.

To enable feasible and practical implementation of the large number of gradient steps to train the models, we use JAX~\cite{jax2018github}, a machine learning framework which has just-in-time (jit) compilation that perform operations on GPU for faster training and can optimize the execution significantly.
We use the QDax library~\cite{lim2022accelerated} to implement this and jit the training of the actor-critic models.

\subsection{Integrating advances in RL within QD-RL}

In this section, we conduct experiments using the Generalized Actor-Critic QD-RL framework to study the benefits of incorporating new advances in Deep RL, namely entropy regularisation (SAC), high C-UTD (DroQ) and critic regularisation (DroQ) within QD-RL.
In our results, we use increased number of critic updates compared to previous literature, following the results from Section~\ref{subsec:train_steps_study}. 
We do so to maximize the performance of each algorithm.

To provide a better intuition on the performance gains obtained from SAC and DroQ over TD3 in the conventional Deep RL setting, we first run these three Deep RL algorithms on the same environments, accounting only for the reward.
Figure~\ref{fig:sac_vs_td3} shows this comparison and corroborate results in literature~\cite{hiraoka2021dropout, smith2022walk}.
We see that SAC performs as well as TD3 on HalfCheetah and Ant.
In the complex Humanoid task, TD3 completely fails while SAC successfully achieves high-performance.
This result shows that both action space exploration coupled with the maximum-entropy objective incentive is crucial to enable Humanoid task to be achievable.

\subsubsection{Benefits of PGA-ME (SAC) - maximum entropy RL} 
\begin{figure}[t!]
\centering
\includegraphics[width =\hsize]{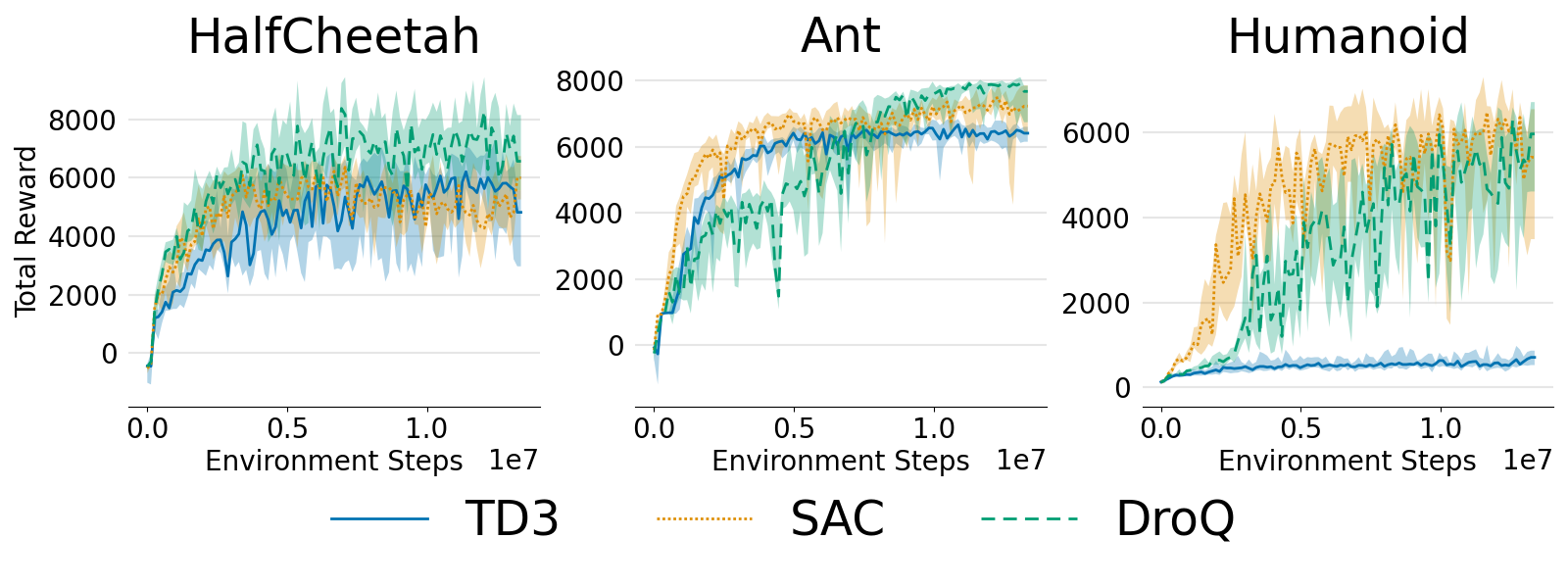}
\vspace{-10pt}\caption{
    Deep RL baselines comparison. Bold lines is the median and shaded areas are the quartiles over 10 replications.
}
\label{fig:sac_vs_td3}
\vspace{-2mm}
\end{figure}

\begin{figure}[t!]
\centering
\includegraphics[width = \hsize]{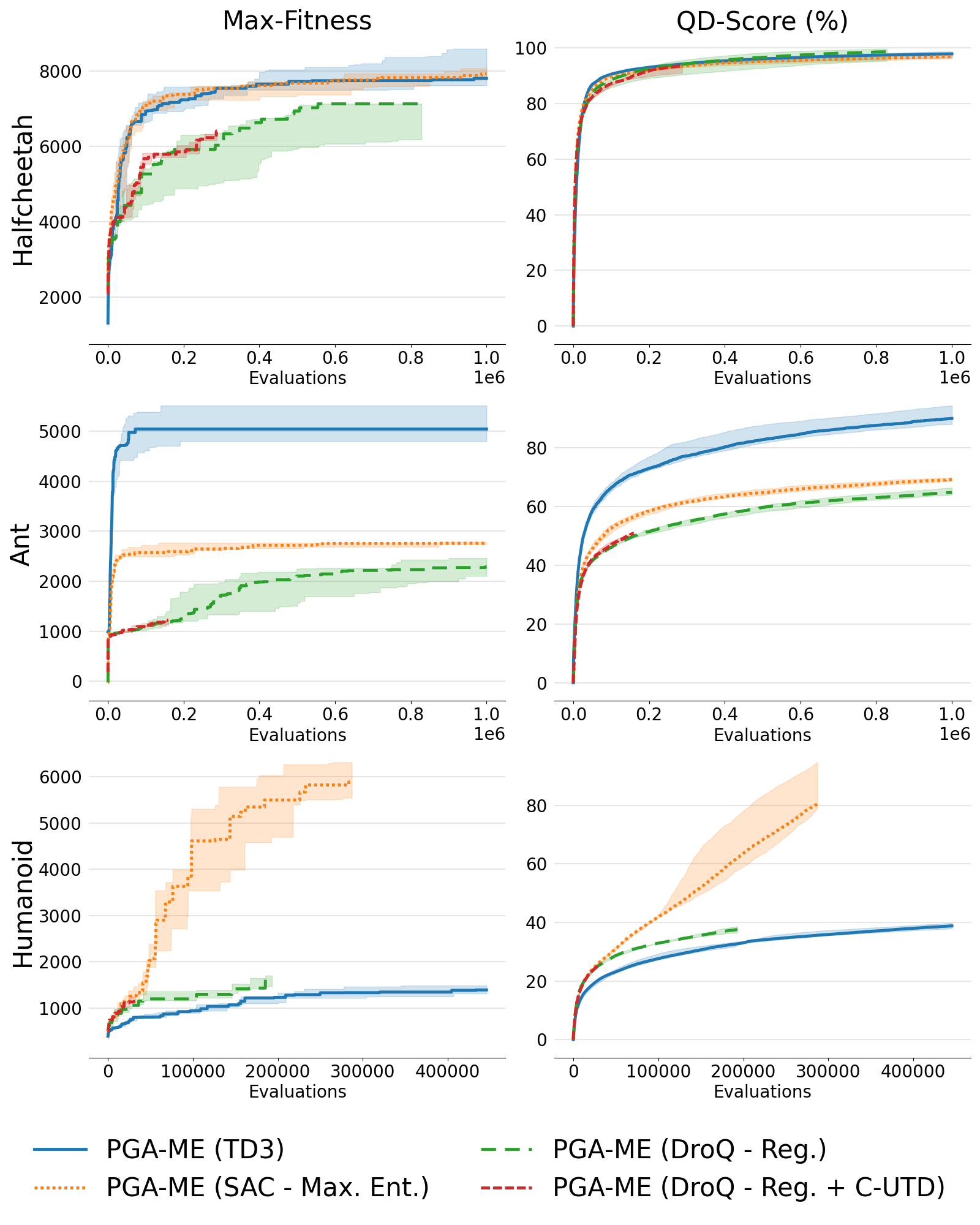}
\vspace{-15pt}\caption{
    Performance of the PGA variants using the TD3, SAC and DroQ RL algorithms formed with the proposed \longname{} framework.
}
\label{fig:pgasac_vs_pgatd3}
\end{figure}

We first evaluate the effectiveness of maximum entropy objectives in the QD-RL setting through PGA-ME (SAC) (Section~\ref{sec:pga_sac}).
Figure~\ref{fig:pgasac_vs_pgatd3} shows the maximum fitness and QD-score metrics.
Most interestingly, we observe that PGA-ME (SAC) is able to achieve in the Humanoid task a maximum fitness that corresponds to solving the task according to previous literature \cite{haarnoja2018soft}. Such scores were never reached by any QD-RL algorithm (including PGA-ME (TD3)) before this.
While Deep RL literature (confirmed by experiments in Figure~\ref{fig:sac_vs_td3}) has already shown that SAC is capable of solving the Humanoid task while TD3 is unable to do so, realizing the same results in the QD-RL setting just by changing the RL algorithm in the framework is a promising indication of the possibilities.
As seen in the Deep RL results, this also demonstrates that action space exploration and maximum-entropy objective of SAC enables the Humanoid task to be solved by PGA-ME (SAC).
We would like to stress here that this results is only permitted by sufficient training of the actor-critic models, as will be shown in Section~\ref{subsec:train_steps_study}.

Results for HalfCheetah similarly follow the Deep RL trend, where PGA-ME (SAC) and PGA-ME (TD3) perform equivalently.
However, on Ant, PGA-ME (SAC) under-performs PGA-ME (TD3) in Max-Fitness and QD-Score ($p<5.10^{-5}$). 
However, when comparing the Max-Fitness with the Total Reward of the Deep RL TD3 baseline, it appears that even PGA-ME (TD3) does not reach the performance of the RL baselines.
Ant has a $4$-dimensional descriptor space, compared to Humanoid and HalfCheetah where the descriptor is $2$-dimensional.
We hypothesise that, due to this higher-dimensionality, the archive contains solutions that differ more than in other tasks. 
This leads to a bigger data distribution shift when moving from Deep RL to QD-RL setting in this task.
Additionally, SAC incentivizes policies that exhibit high entropy resulting in larger diversity of data. Then, it seems that the Update Critic in SAC is more impacted by this shift than the one in TD3.

\subsubsection{Limitations of PGA-ME (DroQ - Reg. + C-UTD)}
Here, we study the increased C-UTD ratio integrated with PGA-ME (DroQ) (see Section~\ref{subsec:critic_training}): 
DroQ uses a C-UTD ratio of $20$, which corresponds to training the critic $20$ times more than in SAC after a transition is collected.
In Generalized Actor-Critic QD-RL, this would sum up to increase the number of critic updates $G_{critic} = 20$ with $G_{actor} = 1$ while keeping the $G$ value chosen for PGA-ME (SAC). 
Here, this would mean doing a total of $\sim 10^5$ critic steps per generations, while PGA-ME (SAC) performs $\sim 10^4$ critic steps per generation. 

This significant increase in amount of training has a major drawback: the time taken to execute the large number of gradient steps. 
Based on the fastest existing open-source DroQ implementation which operates at 2400 gradient steps per second real time on a physical robot~\cite{smith2022walk}, we can estimate that taking $10^6$ critic steps require $7$ minutes on GPU in the best scenario. For a run of $10^6$ evaluations with batch-size $128$, this would thus require almost $900$ hours of training on a GPU, which is clearly impractical. 
We still run PGA-ME (DroQ - Reg. + C-UTD) in Figure~\ref{fig:pgasac_vs_pgatd3}, capping its training time and it appears that performance increase is minimal considering the required budget ($p<5.10^{-3}$). 
This first dimension highlights an interesting point: the difference in settings can make changes in Deep RL impractical for the QD-RL framework.

\subsubsection{Limitations of PGA-ME (DroQ - Reg.)}
The second insight proposed in DroQ is the regularization of the critic in the form of dropout and layer normalisation.
Considering the impracticability and ineffectiveness of PGA-ME (DroQ - Reg. + C-UTD), we evaluate regularization independently from the C-UTD ratio. We name this approach PGA-ME (DroQ - Reg.) in Figure~\ref{fig:pgasac_vs_pgatd3}.

We observe a drop in performance across all environments when using PGA-ME (DroQ  - Reg.) ($p < 5.10^{-3}$).
This result gives another example where insights from Deep RL do not transfer well to QD-RL due to difference in setup.
In DroQ, the goal of the critic regularization is to prevent overfitting of the critic due to the high C-UTD setting.
We hypothesize that this insight does not effectively transfer because in QD-RL, data is collected by a large number of different policies and not a single policy.
In comparison, the transitions seen in Deep RL algorithms are very biased toward the latest actor, despite the off-policy setting.
Hence, overfitting is less likely on the more diverse data distribution resulting from the different policies. 
It appears that further regularizing the critic in this procedure would be detrimental.

\subsection{Importance of sufficient critic and actor training in QD-RL} \label{subsec:train_steps_study}
In this section, we experimentally study the observation obtained from a principled analysis in Section~\ref{subsec:methods_study} that the actor-critic models are usually under-trained in QD-RL.
To do so, we study the impact of modifying the value of $G$, the number of update loops, on the best performing PGA-ME variant for each environment.

\begin{figure}[t!]
\centering
\includegraphics[width = \hsize]{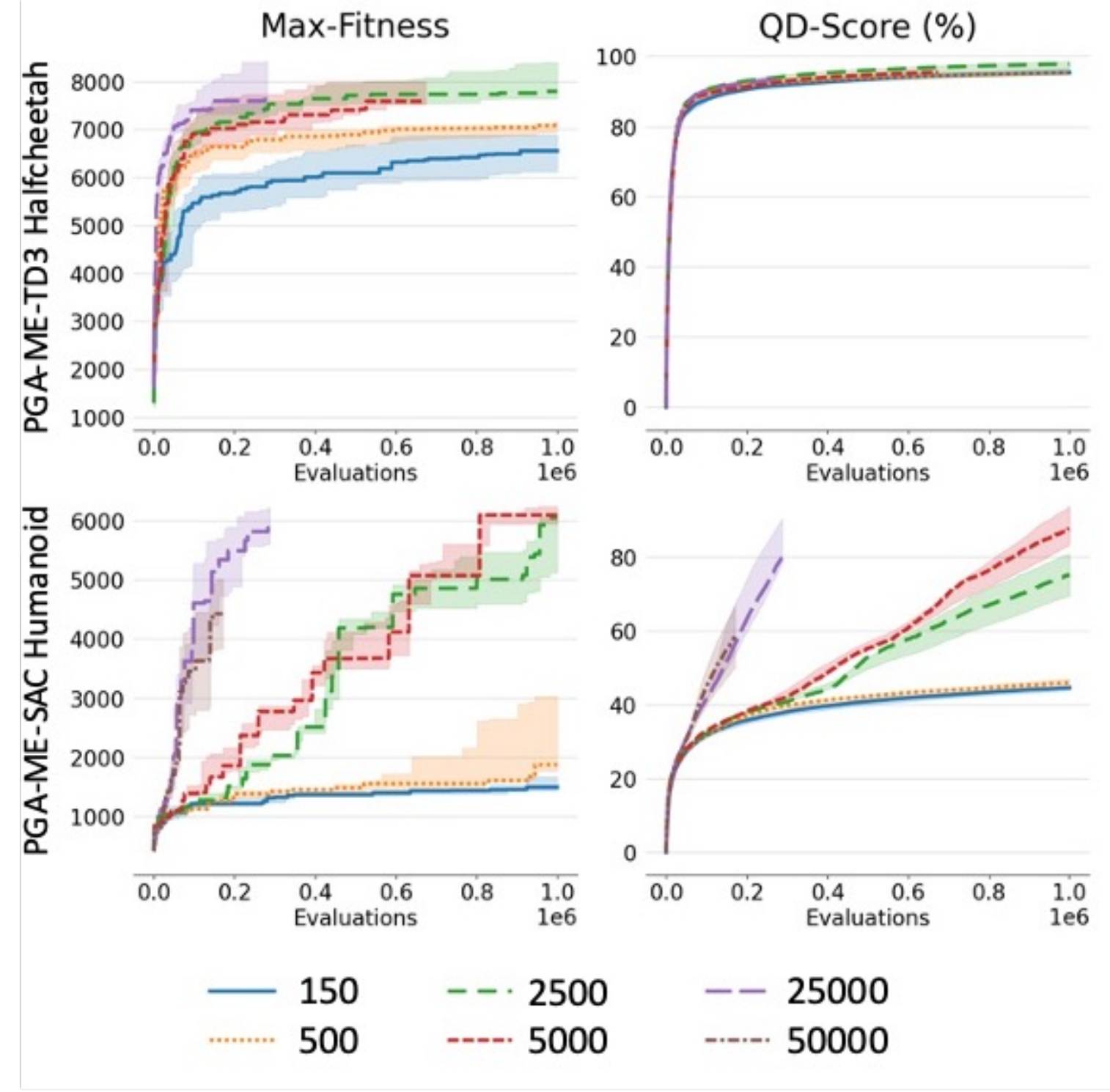}
\vspace{-10pt}\caption{
    Results showing the positive effect of increasing $G$ across both PGA-ME-TD3 and PGA-ME-SAC.
}
\label{fig:critic_steps_results}
\end{figure}

\input{figures/add_table.tex}

\subsubsection{Rationale for value of $G$}
First, we aim to estimate a reasonable range for $G$. 
Given the batch size $B=128$ and episode length $T=1,000$ used, at each generation $128,000$ transitions are added to the replay buffer $\mathcal{D}$.
Using the principled approach discussed in Section~\ref{subsec:methods_study}, if we follow the Deep RL setting in which a gradient step is taken for each transition added to $\mathcal{D}$, the QD-RL setting should similarly take $128,000$ gradient steps after each generation.
We run a study of increasing $G$ from the original author implementation of $G=150$, which corresponds to $300$ critic steps until a maximum number of $G=50,000$, which gives $100,000$ critic steps.

\subsubsection{Effect of increasing $G$}
Despite optimizing the code with jit, the time taken for training the actor-critic models and hence the runtime of the entire algorithm, increases with the update loops $G$ taken at each generation.
For practical implementation, we stop the run of each algorithm after $21$ hours of running.
This results in variants which have large number of gradient steps not reaching the budget number of evaluations. 
Figure~\ref{fig:critic_steps_results} shows the QD-score and Max-fitness curves for these experiments.
We observe that the performance of PGA-ME improves with an increase in the number of update loops $G$. This is true across PGA-ME (TD3) and PGA-ME (SAC).
We also find that it is critical to have a sufficiently large $G$ for PGA-ME (SAC) to be able to solve the Humanoid task.
This indicates that the actor-critic models were severely under-trained as values which are optimal seem to be at least $10\times$ more than used in prior work ($p < 5.10^{-3}$).
Additionally, this increase does not require any additional environment evaluations, leading to an increase in sample efficiency.

\subsubsection{Explaining the effect of increasing $G$ with utility of gradient-directed variations}
To further understand the improvement obtained from increasing $G$, we display in Table~\ref{tab:variation_usage} the variation operators metrics~\cite{flageat2022empirical}. 
This metric corresponds to the number of individuals added to the archive at each generation, by each different variation in the algorithm. 
Here, we give the average utility of the actor and of the offspring induced by gradient-directed (PG) mutations.
We observe a significant increase in both utilities when increasing $G$.
This indicates that increasing $G$ results in better actor and critic.
A better actor can lead to the propagation of this good individual through the mutations~\cite{flageat2022empirical}, while a better critic can improve the PG variations, resulting in an overall better QD score.

\section{Discussion and Conclusion}
In this paper, we introduce Generalized Actor-Critic QD-RL, a modular framework that aims to encourage advances in QD-RL algorithms by facilitating the evaluation of progress in Deep RL within QD-RL algorithms. 
Each module of this framework can easily be swapped out to integrate new ideas and insights from Deep RL. 
Hence, it constitutes a critical and efficient way to facilitate progress in QD-RL.
We demonstrate this generalization by presenting PGA-ME (SAC) and PGA-ME (DroQ), two new algorithms which use this framework to incorporate advances in Deep RL over the original TD3 algorithm used in PGA-ME.
PGA-ME (SAC) enables the Humanoid environment to be solved, which was not possible with existing QD-RL methods.
However, the high C-UTD settings in PGA-ME (DroQ) are less practical in QD-RL due to time constraints while critic regularization, also from PGA-ME (DroQ), does not easily translate from Deep RL to QD. This interestingly highlights that not every advancement in Deep RL can be translated to QD-RL.
Additionally, we also find that the number of training steps of the actor-critic models is critical to the performance of QD-RL algorithms.
Our study shows that these models were under-trained in prior work and can be explained by analyzing the differences between Deep RL and QD settings. 

Overall, we believe Generalized Actor-Critic QD-RL is an important step in understanding the powerful synergies between QD and Deep RL and using them effectively.
Our work demonstrates that not all insights from Deep RL can be effectively transferred and used in the QD-RL setting.
More importantly, a principled analysis of the differences between these two settings can provide reasons behind the effectiveness or ineffectiveness of these insights.

While we show that naive usage of some methods (DroQ - regularization and C-UTD) do not translate directly to QD-RL algorithms, further insights from Deep RL accounting for the off-policy and off-line data could also be used.
Another limitation that is left unexplored in this work is the reproducibility of the policies learnt~\cite{flageat2022empirical}.
As SAC uses stochastic policies during evaluation, instead of deterministic ones as in TD3, they have the potential to behave differently from one evaluation to another. 
We hope to study these more in future work.

\begin{acks}
This work was supported by the Engineering and Physical Sciences Research Council (EPSRC) grant EP/V006673/1 project REcoVER. 
\end{acks}

\bibliographystyle{ACM-Reference-Format}
\bibliography{references}

\newpage

\appendix

\section{Appendix}

\subsection{Supplementary Results}

\subsubsection{Variation}
Table~\ref{tab:variation_usage_full} shows the full variation utility when increasing the parameter $G$.

\begin{table}[h]
\footnotesize
\centering
\begin{tabular}{l|ccccc}
\toprule
\textsc{Algorithm}
 & \multicolumn{5}{c}{\textsc{PGA-TD3 (Half cheetah) }}
\\ 
Num. update loops, $G$
 & \textsc{150}
 & \textsc{500}
 & \textsc{2500} 
  & \textsc{5000}
   & \textsc{25000} 
 \\ 
\midrule
\addlinespace
\textsc{Actor addition}
& 0.032 & 0.034 & 0.405 & 0.509 & 0.917 \\
\textsc{PG addition} 
& 0.244 & 0.247 & 0.377 & 0.432 & 0.746\\
\textsc{GA addition} 
& 2.26 & 2.03 & 1.99 & 2.91 & 5.20 \\
\bottomrule
\end{tabular}
\caption{
        Average addition utility of the actor, PG variations and GA variations with small and large values of $G$, num. of update loops for the Ant task.
}
\label{tab:variation_usage_full}
\end{table}

\begin{table}[h!]
\footnotesize
\centering
\begin{tabular}{l|cccccc}
\toprule
\textsc{Algorithm}
 & \multicolumn{6}{c}{\textsc{PGA-SAC (Humanoid) } }
\\ 
Num. update loops, $G$
 & \textsc{150}
 & \textsc{500}
 & \textsc{2500} 
  & \textsc{5000}
   & \textsc{25000} 
  & \textsc{50000}
 \\ 
\midrule
\addlinespace
\textsc{Actor addition}
& 0.058 & 0.062 & 0.190 & 0.173 & 0.29 & 0.24  \\
\textsc{PG addition} 
& 0.760 & 0.765 & 0.840 & 0.866 & 2.38 & 3.48 \\
\textsc{GA addition} 
& 1.22 & 1.20 & 1.45 & 1.52 & 3.90 & 5.23 \\
\bottomrule
\end{tabular}
\caption{
        Average addition utility of the actor, PG variations and GA variations with increasing values of $G$, num. of update loops for the Humanoid task
}
\label{tab:variation_usage_full}
\end{table}

\subsubsection{Archive Visualizations}
\begin{figure*}[t!]
\centering
\includegraphics[width =\hsize]{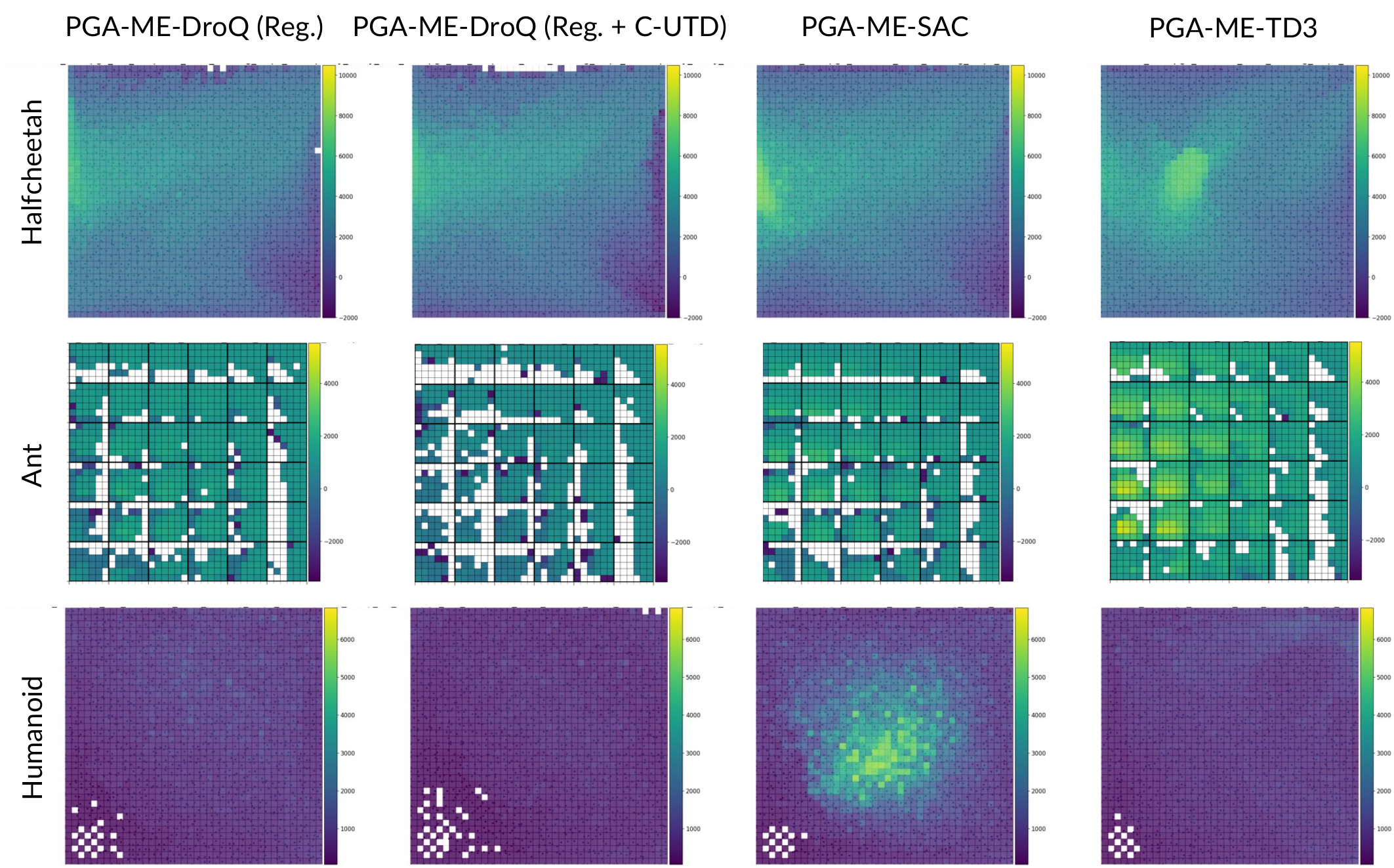}
\caption{
    Archive visualization of the PGA-ME variants (TD3, SAC, DroQ) considered across the different tasks.
}
\label{fig:archive_plots}
\vspace{-2mm}
\end{figure*}

Figure~\ref{fig:archive_plots} show the archive of elites for the different PGA-ME variants with varying Deep RL baseline algorithms enabled by our framework.

\subsection{Experimental and Implementation Details}

\subsubsection{Hyper-parameters}
Table~\ref{tab:hyperparams} shows the list of hyperparameters used for the different PGA-ME variants used. The row blocks are used to differentiate hyperparameters specific to a particular algorithm.

\input{figures/hyperparameters_table.tex}

\end{document}

%% file: notations.tex
\newcommand{\pop}[0]{\Theta}
\newcommand{\addbuffer}[0]{\mathcal{B}_{add}}
\newcommand{\batchsize}[0]{B}
\newcommand{\iter}[0]{j}
\newcommand{\numiterations}[0]{J}
\newcommand{\popit}[0]{\Theta_j}
\newcommand{\maxmodelsteps}[0]{n_{steps}}
\newcommand{\longname}[0]{Generalized Actor-Critic QD-RL}
\newcommand{\name}[0]{GAC}
\newcommand{\PGASAC}[0]{PGA (SAC)}
\newcommand{\PGADroq}[0]{PGA (DroQ - high $G_{critic})$}
\newcommand{\PGAReg}[0]{PGA (DroQ - regularization)}

\newcommand{\numenvsteps}[0]{N}

\newcommand\mycommfont[1]{\small\ttfamily\textcolor{blue}{#1}}

%% file: algorithms/deeprl_critic_training.tex
\SetCommentSty{mycommfont}
\begin{algorithm}[t]
\footnotesize
\caption{Single-Policy Deep RL Training} 
\label{alg:deeprl_critic_train}
    \textbf{Inputs:} num. of critic steps $G_{Q_\psi}$, num. of policy steps $G_{\pi_{\theta_a}}$
    
    \textbf{Init:} critic $Q_{\psi}$, actor $\pi_{\theta_a}$, replay buffer $\mathcal{D}$
    
    \For{$\iter{} = 1, ..., \numenvsteps{}$}{

        \For{$t, ..., T{}$}{
            \tcp{Collect data in the environment}
            $a_t \sim \pi_{\theta_a}(a_t|s_t)$
            
            $s_{t+1} \sim p(s_{t+1}|s_t, a_t)$
    
            $\mathcal{D} \gets (s_t, a_t, s_{t+1}, r_t)$
    
            \tcp{Update critic and actor}
                \For{$g_{Q}= 1, ..., G_{critic}$}{
                    $Q_\psi \gets \text{update\_critic}(Q_\psi, \mathcal{D})$
                }
                \For{$g_{\pi}= 1, ..., G_{actor}$}{
                    $\pi_{\theta_a} \gets \text{update\_actor}(Q_\psi, \pi_{\theta_a}, \mathcal{D})$
                }
        }
    }
\Return $\pi_{\theta_a}$
\end{algorithm}

%% file: algorithms/td3_update.tex
\SetCommentSty{mycommfont}
\SetKwProg{Def}{def}{:}{}
\begin{algorithm}[t]
\footnotesize
\caption{TD3 critic and policy update} 
\label{alg:td3_training}

    \Def{\text{update\_critic}($Q_\psi$, $\mathcal{D}$)}{
        \tcp{Sample $N$ transitions from $\mathcal{D}$}
        $(s_t, a_t, s_{t+1}, r_t) \sim \mathcal{D}$
        
        \tcp{$Q_{\psi'}$ and $\pi_{\theta'}$ represents targets networks}
        $\epsilon \sim \text{clip}(\mathcal{N}(0, \sigma_p), -c, c)$

        $y = r(s_t, a_t, s_{t+1}) + \min_{i=1,2}Q_{\psi'_i}(s_{t+1}, \pi_{\theta'}(s_{t+1}) + \epsilon)$

        $\psi_i \gets \frac{1}{N} \sum{(y - Q_{\psi_i}(s, a))^2}$ 

        \Return $Q_\psi$
    }

    \Def{\text{update\_policy}($Q_\psi$, $\pi_{\theta_a}$, $\mathcal{D}$)}{
        \tcp{Sample $N$ transitions from $\mathcal{D}$}
        $(s_t, a_t, s_{t+1}, r_t) \sim \mathcal{D}$
        
        $\theta_a \gets \frac{1}{N} \sum{Q_{\psi_1}(s_t, a_t)}$

        \Return $\pi_{\theta_a}$
    }

\end{algorithm}

%% file: algorithms/sac_update.tex
\SetCommentSty{mycommfont}
\SetKwProg{Def}{def}{:}{}
\begin{algorithm}[t]
\footnotesize
\caption{SAC critic and policy update} 
\label{alg:sac_training}

    \Def{\text{update\_critic}($Q_\psi$, $\mathcal{D}$)}{
        \tcp{Sample $N$ transitions from $\mathcal{D}$}
        $(s_t, a_t, s_{t+1}, r_t) \sim \mathcal{D}$
        
        \tcp{$Q_{\psi'}$ represents targets networks}
        $y = r_t + \min_{i=1,2}Q_{\psi'_i}(s_{t+1}, \pi_\theta(.|s_{t+1}))$

        $\psi_i \gets \frac{1}{N} \sum{(y - Q_{\psi_i}(s, a))^2}$ 

        \Return $Q_\psi$
    }

    \Def{\text{update\_policy}($Q_\psi$, $\pi_{\theta_a}$, $\mathcal{D}$)}{
        \tcp{Sample $N$ transitions from $\mathcal{D}$}
        $(s_t, a_t, s_{t+1}, r_t) \sim \mathcal{D}$
        
        $\theta_a \gets \frac{1}{N} \sum{(\min_{i=1,2}Q_{\psi}(s_t, a_t) + )}$

        \Return $\pi_{\theta_a}$
    }
   
\end{algorithm}

%% file: algorithms/gac.tex
\SetCommentSty{mycommfont}
\begin{algorithm}[t]
\footnotesize
\caption{Generalized Actor-Critic QD-RL algorithm - \hlgray{highlighted text} indicates the modular parts that can easily be swapped to form new algorithms} 
\label{alg:gac}
    \textbf{Inputs:} num. generations $J$, batch-size $B$
    
    \textbf{Init:} \hlgray{container $\pop$}, \hlgray{data organizer $\mathcal{D}$}, critic $Q_{\psi}$, actor $\pi_{\theta_a}$
    
    \For{$\iter{} = 1, ..., \numiterations{}$}{

        \For{$b = 1, ..., B$}{

            \tcp{Variation}
            $\pi_{\theta_b} \gets$ \hlgray{selector$(\pop)$}
            
            \If{$b > B *$ \hlgray{$p_{PG}$}}{
                $\pi_{\tilde{\theta_b}} \gets$ \hlgray{GA\_variation$(\pi_{\theta_b}, \pop)$}
            }
            \Else{
                \For{$g= 1, ...,$ \hlgray{$G_{PG}$}}{
                    $\pi_{\tilde{\theta_b}} \gets$ \hlgray{$ \text{update\_actor}(\pi_{\theta_b}, Q_\psi, \pi_{\theta_a}, \mathcal{D})$}
                }
            }
            \tcp{Evaluation}
            \For{$t, ..., T{}$}{
                $a_t \sim \pi_{\theta_b}(a_t|s_t)$
                
                $s_{t+1} \sim p(s_{t+1}|s_t, a_t)$

                $\tau_{\tilde{\theta_b}} \gets (s_t, a_t, s_{t+1}, r_t)$
            }
            $F(\pi_{\tilde{\theta_b}}), d(\pi_{\tilde{\theta_b}}) \gets  \text{evaluate}(\tau_{\tilde{\theta_b}})$

            \tcp{Update data organizer and container}
            
            $\mathcal{D} \gets$ \hlgray{update\_data$(\mathcal{D}, \tau_{\tilde{\theta_b}})$}

            $\pop \gets$ \hlgray{update\_container $ (\pop, F(\pi_{\tilde{\theta_b}}), d(\pi_{\tilde{\theta_b}}), \pi_{\tilde{\theta_b}})$}
        }
        
        \tcp{Update critic and actor}
        \For{$g= 1, ...,$ \hlgray{$G$}}{
            \For{$g= 1, ...,$ \hlgray{$G_{critic}$}}{
                $Q_\psi \gets$ \hlgray{update\_critic $(Q_\psi, \mathcal{D})$} 
            }
            \For{$g= 1, ...,$ \hlgray{$G_{actor}$}}{
                $\pi_{\theta_a} \gets$ \hlgray{$ \text{update\_actor}(\pi_{\theta_a}, Q_\psi, \pi_{\theta_a}, \mathcal{D})$}
            }
        }
    }
\Return $\pop$
\end{algorithm}

%% file: figures/add_table.tex
\begin{table}[]
\footnotesize
\centering
\begin{tabular}{l|cc|cc}
\toprule
\textsc{Algorithm}
 & \multicolumn{2}{c}{\textsc{PGA-TD3 (Half cheetah) }}
 & \multicolumn{2}{c}{\textsc{PGA-SAC (Humanoid) } }
\\ 
Num. update loops, $G$
 & \textsc{150}
 & \textsc{2,500}
 & \textsc{150} 
  & \textsc{25,000}
 \\ 
\midrule
\addlinespace
\textsc{Actor addition}
& 0.032 & \textbf{0.041} & 0.058 & \textbf{0.28} \\
\textsc{PG addition} 
& 0.244 & \textbf{0.377} & 0.760 & \textbf{2.38}
\\
\bottomrule
\end{tabular}
\caption{
        Average addition utility of the actor and PG variations with small and large values of $G$, num. of update loops.
}
\label{tab:variation_usage}
\vspace{-4mm}
\end{table}

%% file: figures/hyperparameters_table.tex
\begin{table*}[htb]
\centering
\begin{tabular}{l|ccc}
\toprule
\textsc{Hyperparameter} & \textsc{PGA-ME-TD3} & \textsc{PGA-ME-SAC} & \textsc{\textsc{PGA-ME-DroQ}} \\ 
\midrule
\addlinespace
Policy hidden layer sizes & [64, 64] & [64, 64] & [64, 64] \\
Batch size, $\batchsize{}$ & 128 & 128 & 128\\
\midrule
\addlinespace
Iso coefficient, $\sigma_1$  & 0.01 & 0.01 & 0.01 \\
Line coefficient, $\sigma_2$ & 0.1 & 0.1 & 0.1 \\
\midrule
\addlinespace
Proportion PG $p_{PG}$ &  0.5 & 0.5 &  0.5 \\ 
Num. PG training steps $G_{PG}$ &  100 &  100 &  100  \\ 
PGA replay buffer size &  $10^6$ & $10^6$ & $10^6$ \\
Critic hidden layer size & [256, 256] &  [256, 256] & [256, 256] \\
Critic learning rate & 0.0003 & 0.0003 &  0.0003   \\ 
Greedy learning rate & 0.0003  & 0.0003 & 0.0003      \\ 
PG variation learning rate & 0.001 &  0.001 & 0.001 \\
Transitions batch size & 256 &  256 & 256 \\
\midrule
\addlinespace
Noise clip & 0.5  &  - & - \\
Policy noise & 0.2 &  - & - \\
Soft $\tau$ update & 0.005 &  - & - \\
\midrule
\addlinespace
Entropy coefficient init $\alpha_{init}$ & - &  1.0 & 1.0 \\
$\tau$ update & - & 0.005 & 0.005 \\
\midrule
\addlinespace
Dropout rate & - & - & 0.01 \\
\bottomrule
\end{tabular}
\caption{
    Hyperparameters of PGA-ME variants - TD3, SAC and DroQ.
}
\label{tab:hyperparams}
\end{table*}

%% file: main.bbl

\begin{thebibliography}{38}


\ifx \showCODEN    \undefined \def \showCODEN     #1{\unskip}     \fi
\ifx \showDOI      \undefined \def \showDOI       #1{#1}\fi
\ifx \showISBNx    \undefined \def \showISBNx     #1{\unskip}     \fi
\ifx \showISBNxiii \undefined \def \showISBNxiii  #1{\unskip}     \fi
\ifx \showISSN     \undefined \def \showISSN      #1{\unskip}     \fi
\ifx \showLCCN     \undefined \def \showLCCN      #1{\unskip}     \fi
\ifx \shownote     \undefined \def \shownote      #1{#1}          \fi
\ifx \showarticletitle \undefined \def \showarticletitle #1{#1}   \fi
\ifx \showURL      \undefined \def \showURL       {\relax}        \fi
\providecommand\bibfield[2]{#2}
\providecommand\bibinfo[2]{#2}
\providecommand\natexlab[1]{#1}
\providecommand\showeprint[2][]{arXiv:#2}

\bibitem[\protect\citeauthoryear{Akkaya, Andrychowicz, Chociej, Litwin, McGrew,
  Petron, Paino, Plappert, Powell, Ribas, et~al\mbox{.}}{Akkaya
  et~al\mbox{.}}{2019}]%
        {akkaya2019solving}
\bibfield{author}{\bibinfo{person}{Ilge Akkaya}, \bibinfo{person}{Marcin
  Andrychowicz}, \bibinfo{person}{Maciek Chociej}, \bibinfo{person}{Mateusz
  Litwin}, \bibinfo{person}{Bob McGrew}, \bibinfo{person}{Arthur Petron},
  \bibinfo{person}{Alex Paino}, \bibinfo{person}{Matthias Plappert},
  \bibinfo{person}{Glenn Powell}, \bibinfo{person}{Raphael Ribas},
  {et~al\mbox{.}}} \bibinfo{year}{2019}\natexlab{}.
\newblock \showarticletitle{Solving rubik's cube with a robot hand}.
\newblock \bibinfo{journal}{\emph{arXiv preprint arXiv:1910.07113}}
  (\bibinfo{year}{2019}).
\newblock


\bibitem[\protect\citeauthoryear{Bradbury, Frostig, Hawkins, Johnson, Leary,
  Maclaurin, Necula, Paszke, Vander{P}las, Wanderman-{M}ilne, and
  Zhang}{Bradbury et~al\mbox{.}}{2018}]%
        {jax2018github}
\bibfield{author}{\bibinfo{person}{James Bradbury}, \bibinfo{person}{Roy
  Frostig}, \bibinfo{person}{Peter Hawkins}, \bibinfo{person}{Matthew~James
  Johnson}, \bibinfo{person}{Chris Leary}, \bibinfo{person}{Dougal Maclaurin},
  \bibinfo{person}{George Necula}, \bibinfo{person}{Adam Paszke},
  \bibinfo{person}{Jake Vander{P}las}, \bibinfo{person}{Skye
  Wanderman-{M}ilne}, {and} \bibinfo{person}{Qiao Zhang}.}
  \bibinfo{year}{2018}\natexlab{}.
\newblock \bibinfo{booktitle}{\emph{{JAX}: composable transformations of
  {P}ython+{N}um{P}y programs}}.
\newblock
\urldef\tempurl%
\url{http://github.com/google/jax}
\showURL{%
\tempurl}


\bibitem[\protect\citeauthoryear{Chalumeau, Boige, Lim, Mac{\'e}, Allard,
  Flajolet, Cully, and Pierrot}{Chalumeau et~al\mbox{.}}{2022}]%
        {chalumeau2022neuroevolution}
\bibfield{author}{\bibinfo{person}{Felix Chalumeau}, \bibinfo{person}{Raphael
  Boige}, \bibinfo{person}{Bryan Lim}, \bibinfo{person}{Valentin Mac{\'e}},
  \bibinfo{person}{Maxime Allard}, \bibinfo{person}{Arthur Flajolet},
  \bibinfo{person}{Antoine Cully}, {and} \bibinfo{person}{Thomas Pierrot}.}
  \bibinfo{year}{2022}\natexlab{}.
\newblock \showarticletitle{Neuroevolution is a Competitive Alternative to
  Reinforcement Learning for Skill Discovery}.
\newblock \bibinfo{journal}{\emph{arXiv preprint arXiv:2210.03516}}
  (\bibinfo{year}{2022}).
\newblock


\bibitem[\protect\citeauthoryear{Chatzilygeroudis, Cully, Vassiliades, and
  Mouret}{Chatzilygeroudis et~al\mbox{.}}{2021}]%
        {chatzilygeroudis2021quality}
\bibfield{author}{\bibinfo{person}{Konstantinos Chatzilygeroudis},
  \bibinfo{person}{Antoine Cully}, \bibinfo{person}{Vassilis Vassiliades},
  {and} \bibinfo{person}{Jean-Baptiste Mouret}.}
  \bibinfo{year}{2021}\natexlab{}.
\newblock \showarticletitle{Quality-Diversity Optimization: a novel branch of
  stochastic optimization}.
\newblock In \bibinfo{booktitle}{\emph{Black Box Optimization, Machine
  Learning, and No-Free Lunch Theorems}}. \bibinfo{publisher}{Springer},
  \bibinfo{pages}{109--135}.
\newblock


\bibitem[\protect\citeauthoryear{Chen, Wang, Zhou, and Ross}{Chen
  et~al\mbox{.}}{2021}]%
        {chen2021randomized}
\bibfield{author}{\bibinfo{person}{Xinyue Chen}, \bibinfo{person}{Che Wang},
  \bibinfo{person}{Zijian Zhou}, {and} \bibinfo{person}{Keith Ross}.}
  \bibinfo{year}{2021}\natexlab{}.
\newblock \showarticletitle{Randomized ensembled double q-learning: Learning
  fast without a model}.
\newblock \bibinfo{journal}{\emph{arXiv preprint arXiv:2101.05982}}
  (\bibinfo{year}{2021}).
\newblock


\bibitem[\protect\citeauthoryear{Colas, Madhavan, Huizinga, and Clune}{Colas
  et~al\mbox{.}}{2020}]%
        {colas2020scaling}
\bibfield{author}{\bibinfo{person}{C{\'e}dric Colas}, \bibinfo{person}{Vashisht
  Madhavan}, \bibinfo{person}{Joost Huizinga}, {and} \bibinfo{person}{Jeff
  Clune}.} \bibinfo{year}{2020}\natexlab{}.
\newblock \showarticletitle{Scaling map-elites to deep neuroevolution}. In
  \bibinfo{booktitle}{\emph{Proceedings of the 2020 Genetic and Evolutionary
  Computation Conference}}. \bibinfo{pages}{67--75}.
\newblock


\bibitem[\protect\citeauthoryear{Cully, Clune, Tarapore, and Mouret}{Cully
  et~al\mbox{.}}{2015}]%
        {cully2015robots}
\bibfield{author}{\bibinfo{person}{Antoine Cully}, \bibinfo{person}{Jeff
  Clune}, \bibinfo{person}{Danesh Tarapore}, {and}
  \bibinfo{person}{Jean-Baptiste Mouret}.} \bibinfo{year}{2015}\natexlab{}.
\newblock \showarticletitle{Robots that can adapt like animals}.
\newblock \bibinfo{journal}{\emph{Nature}} \bibinfo{volume}{521},
  \bibinfo{number}{7553} (\bibinfo{year}{2015}), \bibinfo{pages}{503--507}.
\newblock


\bibitem[\protect\citeauthoryear{Cully and Demiris}{Cully and Demiris}{2017}]%
        {cully2017quality}
\bibfield{author}{\bibinfo{person}{Antoine Cully} {and}
  \bibinfo{person}{Yiannis Demiris}.} \bibinfo{year}{2017}\natexlab{}.
\newblock \showarticletitle{Quality and diversity optimization: A unifying
  modular framework}.
\newblock \bibinfo{journal}{\emph{IEEE Transactions on Evolutionary
  Computation}} \bibinfo{volume}{22}, \bibinfo{number}{2}
  (\bibinfo{year}{2017}), \bibinfo{pages}{245--259}.
\newblock


\bibitem[\protect\citeauthoryear{Degrave, Felici, Buchli, Neunert, Tracey,
  Carpanese, Ewalds, Hafner, Abdolmaleki, de~Las~Casas, et~al\mbox{.}}{Degrave
  et~al\mbox{.}}{2022}]%
        {degrave2022magnetic}
\bibfield{author}{\bibinfo{person}{Jonas Degrave}, \bibinfo{person}{Federico
  Felici}, \bibinfo{person}{Jonas Buchli}, \bibinfo{person}{Michael Neunert},
  \bibinfo{person}{Brendan Tracey}, \bibinfo{person}{Francesco Carpanese},
  \bibinfo{person}{Timo Ewalds}, \bibinfo{person}{Roland Hafner},
  \bibinfo{person}{Abbas Abdolmaleki}, \bibinfo{person}{Diego de Las~Casas},
  {et~al\mbox{.}}} \bibinfo{year}{2022}\natexlab{}.
\newblock \showarticletitle{Magnetic control of tokamak plasmas through deep
  reinforcement learning}.
\newblock \bibinfo{journal}{\emph{Nature}} \bibinfo{volume}{602},
  \bibinfo{number}{7897} (\bibinfo{year}{2022}), \bibinfo{pages}{414--419}.
\newblock


\bibitem[\protect\citeauthoryear{Ecoffet, Huizinga, Lehman, Stanley, and
  Clune}{Ecoffet et~al\mbox{.}}{2021}]%
        {ecoffet2021first}
\bibfield{author}{\bibinfo{person}{Adrien Ecoffet}, \bibinfo{person}{Joost
  Huizinga}, \bibinfo{person}{Joel Lehman}, \bibinfo{person}{Kenneth~O
  Stanley}, {and} \bibinfo{person}{Jeff Clune}.}
  \bibinfo{year}{2021}\natexlab{}.
\newblock \showarticletitle{First return, then explore}.
\newblock \bibinfo{journal}{\emph{Nature}} \bibinfo{volume}{590},
  \bibinfo{number}{7847} (\bibinfo{year}{2021}), \bibinfo{pages}{580--586}.
\newblock


\bibitem[\protect\citeauthoryear{Fawzi, Balog, Huang, Hubert, Romera-Paredes,
  Barekatain, Novikov, R~Ruiz, Schrittwieser, Swirszcz, et~al\mbox{.}}{Fawzi
  et~al\mbox{.}}{2022}]%
        {fawzi2022discovering}
\bibfield{author}{\bibinfo{person}{Alhussein Fawzi}, \bibinfo{person}{Matej
  Balog}, \bibinfo{person}{Aja Huang}, \bibinfo{person}{Thomas Hubert},
  \bibinfo{person}{Bernardino Romera-Paredes}, \bibinfo{person}{Mohammadamin
  Barekatain}, \bibinfo{person}{Alexander Novikov},
  \bibinfo{person}{Francisco~J R~Ruiz}, \bibinfo{person}{Julian Schrittwieser},
  \bibinfo{person}{Grzegorz Swirszcz}, {et~al\mbox{.}}}
  \bibinfo{year}{2022}\natexlab{}.
\newblock \showarticletitle{Discovering faster matrix multiplication algorithms
  with reinforcement learning}.
\newblock \bibinfo{journal}{\emph{Nature}} \bibinfo{volume}{610},
  \bibinfo{number}{7930} (\bibinfo{year}{2022}), \bibinfo{pages}{47--53}.
\newblock


\bibitem[\protect\citeauthoryear{Flageat, Chalumeau, and Cully}{Flageat
  et~al\mbox{.}}{2022a}]%
        {flageat2022empirical}
\bibfield{author}{\bibinfo{person}{Manon Flageat}, \bibinfo{person}{Felix
  Chalumeau}, {and} \bibinfo{person}{Antoine Cully}.}
  \bibinfo{year}{2022}\natexlab{a}.
\newblock \showarticletitle{Empirical analysis of PGA-MAP-Elites for
  Neuroevolution in Uncertain Domains}.
\newblock \bibinfo{journal}{\emph{ACM Transactions on Evolutionary Learning}}
  (\bibinfo{year}{2022}).
\newblock


\bibitem[\protect\citeauthoryear{Flageat, Lim, Grillotti, Allard, Smith, and
  Cully}{Flageat et~al\mbox{.}}{2022b}]%
        {flageat2022benchmarking}
\bibfield{author}{\bibinfo{person}{Manon Flageat}, \bibinfo{person}{Bryan Lim},
  \bibinfo{person}{Luca Grillotti}, \bibinfo{person}{Maxime Allard},
  \bibinfo{person}{Simón~C. Smith}, {and} \bibinfo{person}{Antoine Cully}.}
  \bibinfo{year}{2022}\natexlab{b}.
\newblock \bibinfo{title}{Benchmarking Quality-Diversity Algorithms on
  Neuroevolution for Reinforcement Learning}.
\newblock
\newblock
\urldef\tempurl%
\url{https://doi.org/10.48550/ARXIV.2211.02193}
\showDOI{\tempurl}


\bibitem[\protect\citeauthoryear{Freeman, Frey, Raichuk, Girgin, Mordatch, and
  Bachem}{Freeman et~al\mbox{.}}{2021}]%
        {brax2021github}
\bibfield{author}{\bibinfo{person}{C.~Daniel Freeman}, \bibinfo{person}{Erik
  Frey}, \bibinfo{person}{Anton Raichuk}, \bibinfo{person}{Sertan Girgin},
  \bibinfo{person}{Igor Mordatch}, {and} \bibinfo{person}{Olivier Bachem}.}
  \bibinfo{year}{2021}\natexlab{}.
\newblock \bibinfo{booktitle}{\emph{Brax - A Differentiable Physics Engine for
  Large Scale Rigid Body Simulation}}.
\newblock
\urldef\tempurl%
\url{http://github.com/google/brax}
\showURL{%
\tempurl}


\bibitem[\protect\citeauthoryear{Fujimoto, Hoof, and Meger}{Fujimoto
  et~al\mbox{.}}{2018}]%
        {fujimoto2018addressing}
\bibfield{author}{\bibinfo{person}{Scott Fujimoto}, \bibinfo{person}{Herke
  Hoof}, {and} \bibinfo{person}{David Meger}.} \bibinfo{year}{2018}\natexlab{}.
\newblock \showarticletitle{Addressing function approximation error in
  actor-critic methods}. In \bibinfo{booktitle}{\emph{International conference
  on machine learning}}. PMLR, \bibinfo{pages}{1587--1596}.
\newblock


\bibitem[\protect\citeauthoryear{Gaier, Asteroth, and Mouret}{Gaier
  et~al\mbox{.}}{2018}]%
        {gaier2018data}
\bibfield{author}{\bibinfo{person}{Adam Gaier}, \bibinfo{person}{Alexander
  Asteroth}, {and} \bibinfo{person}{Jean-Baptiste Mouret}.}
  \bibinfo{year}{2018}\natexlab{}.
\newblock \showarticletitle{Data-efficient design exploration through
  surrogate-assisted illumination}.
\newblock \bibinfo{journal}{\emph{Evolutionary computation}}
  \bibinfo{volume}{26}, \bibinfo{number}{3} (\bibinfo{year}{2018}),
  \bibinfo{pages}{381--410}.
\newblock


\bibitem[\protect\citeauthoryear{Gravina, Khalifa, Liapis, Togelius, and
  Yannakakis}{Gravina et~al\mbox{.}}{2019}]%
        {gravina2019procedural}
\bibfield{author}{\bibinfo{person}{Daniele Gravina}, \bibinfo{person}{Ahmed
  Khalifa}, \bibinfo{person}{Antonios Liapis}, \bibinfo{person}{Julian
  Togelius}, {and} \bibinfo{person}{Georgios~N Yannakakis}.}
  \bibinfo{year}{2019}\natexlab{}.
\newblock \showarticletitle{Procedural content generation through quality
  diversity}. In \bibinfo{booktitle}{\emph{2019 IEEE Conference on Games
  (CoG)}}. IEEE, \bibinfo{pages}{1--8}.
\newblock


\bibitem[\protect\citeauthoryear{Haarnoja, Zhou, Abbeel, and Levine}{Haarnoja
  et~al\mbox{.}}{2018}]%
        {haarnoja2018soft}
\bibfield{author}{\bibinfo{person}{Tuomas Haarnoja}, \bibinfo{person}{Aurick
  Zhou}, \bibinfo{person}{Pieter Abbeel}, {and} \bibinfo{person}{Sergey
  Levine}.} \bibinfo{year}{2018}\natexlab{}.
\newblock \showarticletitle{Soft actor-critic: Off-policy maximum entropy deep
  reinforcement learning with a stochastic actor}. In
  \bibinfo{booktitle}{\emph{International conference on machine learning}}.
  PMLR, \bibinfo{pages}{1861--1870}.
\newblock


\bibitem[\protect\citeauthoryear{Hiraoka, Imagawa, Hashimoto, Onishi, and
  Tsuruoka}{Hiraoka et~al\mbox{.}}{2021}]%
        {hiraoka2021dropout}
\bibfield{author}{\bibinfo{person}{Takuya Hiraoka}, \bibinfo{person}{Takahisa
  Imagawa}, \bibinfo{person}{Taisei Hashimoto}, \bibinfo{person}{Takashi
  Onishi}, {and} \bibinfo{person}{Yoshimasa Tsuruoka}.}
  \bibinfo{year}{2021}\natexlab{}.
\newblock \showarticletitle{Dropout Q-Functions for Doubly Efficient
  Reinforcement Learning}.
\newblock \bibinfo{journal}{\emph{arXiv preprint arXiv:2110.02034}}
  (\bibinfo{year}{2021}).
\newblock


\bibitem[\protect\citeauthoryear{Jiang, Salley, Sharma, Keenan, Mullin, and
  Cronin}{Jiang et~al\mbox{.}}{2022}]%
        {jiang2022chemical}
\bibfield{author}{\bibinfo{person}{Yibin Jiang}, \bibinfo{person}{Daniel
  Salley}, \bibinfo{person}{Abhishek Sharma}, \bibinfo{person}{Graham Keenan},
  \bibinfo{person}{Margaret Mullin}, {and} \bibinfo{person}{Leroy Cronin}.}
  \bibinfo{year}{2022}\natexlab{}.
\newblock \showarticletitle{An artificial intelligence enabled chemical
  synthesis robot for exploration and optimization of nanomaterials}.
\newblock \bibinfo{journal}{\emph{Science Advances}} \bibinfo{volume}{8},
  \bibinfo{number}{40} (\bibinfo{year}{2022}), \bibinfo{pages}{eabo2626}.
\newblock
\urldef\tempurl%
\url{https://doi.org/10.1126/sciadv.abo2626}
\showDOI{\tempurl}
\showeprint{https://www.science.org/doi/pdf/10.1126/sciadv.abo2626}


\bibitem[\protect\citeauthoryear{Kumar, Zhou, Tucker, and Levine}{Kumar
  et~al\mbox{.}}{2020}]%
        {kumar2020conservative}
\bibfield{author}{\bibinfo{person}{Aviral Kumar}, \bibinfo{person}{Aurick
  Zhou}, \bibinfo{person}{George Tucker}, {and} \bibinfo{person}{Sergey
  Levine}.} \bibinfo{year}{2020}\natexlab{}.
\newblock \showarticletitle{Conservative q-learning for offline reinforcement
  learning}.
\newblock \bibinfo{journal}{\emph{Advances in Neural Information Processing
  Systems}}  \bibinfo{volume}{33} (\bibinfo{year}{2020}),
  \bibinfo{pages}{1179--1191}.
\newblock


\bibitem[\protect\citeauthoryear{Lee, Hwangbo, Wellhausen, Koltun, and
  Hutter}{Lee et~al\mbox{.}}{2020}]%
        {lee2020learning}
\bibfield{author}{\bibinfo{person}{Joonho Lee}, \bibinfo{person}{Jemin
  Hwangbo}, \bibinfo{person}{Lorenz Wellhausen}, \bibinfo{person}{Vladlen
  Koltun}, {and} \bibinfo{person}{Marco Hutter}.}
  \bibinfo{year}{2020}\natexlab{}.
\newblock \showarticletitle{Learning quadrupedal locomotion over challenging
  terrain}.
\newblock \bibinfo{journal}{\emph{Science robotics}} \bibinfo{volume}{5},
  \bibinfo{number}{47} (\bibinfo{year}{2020}), \bibinfo{pages}{eabc5986}.
\newblock


\bibitem[\protect\citeauthoryear{Lim, Allard, Grillotti, and Cully}{Lim
  et~al\mbox{.}}{2022}]%
        {lim2022accelerated}
\bibfield{author}{\bibinfo{person}{Bryan Lim}, \bibinfo{person}{Maxime Allard},
  \bibinfo{person}{Luca Grillotti}, {and} \bibinfo{person}{Antoine Cully}.}
  \bibinfo{year}{2022}\natexlab{}.
\newblock \showarticletitle{Accelerated Quality-Diversity for Robotics through
  Massive Parallelism}.
\newblock \bibinfo{journal}{\emph{arXiv preprint arXiv:2202.01258}}
  (\bibinfo{year}{2022}).
\newblock


\bibitem[\protect\citeauthoryear{Mnih, Kavukcuoglu, Silver, Rusu, Veness,
  Bellemare, Graves, Riedmiller, Fidjeland, Ostrovski, et~al\mbox{.}}{Mnih
  et~al\mbox{.}}{2015}]%
        {mnih2015human}
\bibfield{author}{\bibinfo{person}{Volodymyr Mnih}, \bibinfo{person}{Koray
  Kavukcuoglu}, \bibinfo{person}{David Silver}, \bibinfo{person}{Andrei~A
  Rusu}, \bibinfo{person}{Joel Veness}, \bibinfo{person}{Marc~G Bellemare},
  \bibinfo{person}{Alex Graves}, \bibinfo{person}{Martin Riedmiller},
  \bibinfo{person}{Andreas~K Fidjeland}, \bibinfo{person}{Georg Ostrovski},
  {et~al\mbox{.}}} \bibinfo{year}{2015}\natexlab{}.
\newblock \showarticletitle{Human-level control through deep reinforcement
  learning}.
\newblock \bibinfo{journal}{\emph{nature}} \bibinfo{volume}{518},
  \bibinfo{number}{7540} (\bibinfo{year}{2015}), \bibinfo{pages}{529--533}.
\newblock


\bibitem[\protect\citeauthoryear{Mouret and Clune}{Mouret and Clune}{2015}]%
        {mouret2015illuminating}
\bibfield{author}{\bibinfo{person}{Jean-Baptiste Mouret} {and}
  \bibinfo{person}{Jeff Clune}.} \bibinfo{year}{2015}\natexlab{}.
\newblock \showarticletitle{Illuminating search spaces by mapping elites}.
\newblock \bibinfo{journal}{\emph{arXiv preprint arXiv:1504.04909}}
  (\bibinfo{year}{2015}).
\newblock


\bibitem[\protect\citeauthoryear{Nilsson and Cully}{Nilsson and Cully}{2021}]%
        {nilsson2021policy}
\bibfield{author}{\bibinfo{person}{Olle Nilsson} {and} \bibinfo{person}{Antoine
  Cully}.} \bibinfo{year}{2021}\natexlab{}.
\newblock \showarticletitle{Policy gradient assisted MAP-Elites}. In
  \bibinfo{booktitle}{\emph{Proceedings of the Genetic and Evolutionary
  Computation Conference}}. \bibinfo{pages}{866--875}.
\newblock


\bibitem[\protect\citeauthoryear{Pierrot, Mac{\'e}, Chalumeau, Flajolet,
  Cideron, Beguir, Cully, Sigaud, and Perrin-Gilbert}{Pierrot
  et~al\mbox{.}}{2022}]%
        {pierrot2022diversity}
\bibfield{author}{\bibinfo{person}{Thomas Pierrot}, \bibinfo{person}{Valentin
  Mac{\'e}}, \bibinfo{person}{Felix Chalumeau}, \bibinfo{person}{Arthur
  Flajolet}, \bibinfo{person}{Geoffrey Cideron}, \bibinfo{person}{Karim
  Beguir}, \bibinfo{person}{Antoine Cully}, \bibinfo{person}{Olivier Sigaud},
  {and} \bibinfo{person}{Nicolas Perrin-Gilbert}.}
  \bibinfo{year}{2022}\natexlab{}.
\newblock \showarticletitle{Diversity policy gradient for sample efficient
  quality-diversity optimization}. In \bibinfo{booktitle}{\emph{Proceedings of
  the Genetic and Evolutionary Computation Conference}}.
  \bibinfo{pages}{1075--1083}.
\newblock


\bibitem[\protect\citeauthoryear{Pugh, Soros, and Stanley}{Pugh
  et~al\mbox{.}}{2016}]%
        {pugh2016quality}
\bibfield{author}{\bibinfo{person}{Justin~K Pugh}, \bibinfo{person}{Lisa~B
  Soros}, {and} \bibinfo{person}{Kenneth~O Stanley}.}
  \bibinfo{year}{2016}\natexlab{}.
\newblock \showarticletitle{Quality diversity: A new frontier for evolutionary
  computation}.
\newblock \bibinfo{journal}{\emph{Frontiers in Robotics and AI}}
  \bibinfo{volume}{3} (\bibinfo{year}{2016}), \bibinfo{pages}{40}.
\newblock


\bibitem[\protect\citeauthoryear{Schulman, Wolski, Dhariwal, Radford, and
  Klimov}{Schulman et~al\mbox{.}}{2017}]%
        {schulman2017proximal}
\bibfield{author}{\bibinfo{person}{John Schulman}, \bibinfo{person}{Filip
  Wolski}, \bibinfo{person}{Prafulla Dhariwal}, \bibinfo{person}{Alec Radford},
  {and} \bibinfo{person}{Oleg Klimov}.} \bibinfo{year}{2017}\natexlab{}.
\newblock \showarticletitle{Proximal policy optimization algorithms}.
\newblock \bibinfo{journal}{\emph{arXiv preprint arXiv:1707.06347}}
  (\bibinfo{year}{2017}).
\newblock


\bibitem[\protect\citeauthoryear{Silver, Huang, Maddison, Guez, Sifre, Van
  Den~Driessche, Schrittwieser, Antonoglou, Panneershelvam, Lanctot,
  et~al\mbox{.}}{Silver et~al\mbox{.}}{2016}]%
        {silver2016mastering}
\bibfield{author}{\bibinfo{person}{David Silver}, \bibinfo{person}{Aja Huang},
  \bibinfo{person}{Chris~J Maddison}, \bibinfo{person}{Arthur Guez},
  \bibinfo{person}{Laurent Sifre}, \bibinfo{person}{George Van Den~Driessche},
  \bibinfo{person}{Julian Schrittwieser}, \bibinfo{person}{Ioannis Antonoglou},
  \bibinfo{person}{Veda Panneershelvam}, \bibinfo{person}{Marc Lanctot},
  {et~al\mbox{.}}} \bibinfo{year}{2016}\natexlab{}.
\newblock \showarticletitle{Mastering the game of Go with deep neural networks
  and tree search}.
\newblock \bibinfo{journal}{\emph{nature}} \bibinfo{volume}{529},
  \bibinfo{number}{7587} (\bibinfo{year}{2016}), \bibinfo{pages}{484--489}.
\newblock


\bibitem[\protect\citeauthoryear{Smith, Kostrikov, and Levine}{Smith
  et~al\mbox{.}}{2022}]%
        {smith2022walk}
\bibfield{author}{\bibinfo{person}{Laura Smith}, \bibinfo{person}{Ilya
  Kostrikov}, {and} \bibinfo{person}{Sergey Levine}.}
  \bibinfo{year}{2022}\natexlab{}.
\newblock \showarticletitle{A walk in the park: Learning to walk in 20 minutes
  with model-free reinforcement learning}.
\newblock \bibinfo{journal}{\emph{arXiv preprint arXiv:2208.07860}}
  (\bibinfo{year}{2022}).
\newblock


\bibitem[\protect\citeauthoryear{Sutton and Barto}{Sutton and Barto}{2018}]%
        {sutton2018reinforcement}
\bibfield{author}{\bibinfo{person}{Richard~S Sutton} {and}
  \bibinfo{person}{Andrew~G Barto}.} \bibinfo{year}{2018}\natexlab{}.
\newblock \bibinfo{booktitle}{\emph{Reinforcement learning: An introduction}}.
\newblock \bibinfo{publisher}{MIT press}.
\newblock


\bibitem[\protect\citeauthoryear{Tassa, Doron, Muldal, Erez, Li, Casas, Budden,
  Abdolmaleki, Merel, Lefrancq, et~al\mbox{.}}{Tassa et~al\mbox{.}}{2018}]%
        {tassa2018deepmind}
\bibfield{author}{\bibinfo{person}{Yuval Tassa}, \bibinfo{person}{Yotam Doron},
  \bibinfo{person}{Alistair Muldal}, \bibinfo{person}{Tom Erez},
  \bibinfo{person}{Yazhe Li}, \bibinfo{person}{Diego de~Las Casas},
  \bibinfo{person}{David Budden}, \bibinfo{person}{Abbas Abdolmaleki},
  \bibinfo{person}{Josh Merel}, \bibinfo{person}{Andrew Lefrancq},
  {et~al\mbox{.}}} \bibinfo{year}{2018}\natexlab{}.
\newblock \showarticletitle{Deepmind control suite}.
\newblock \bibinfo{journal}{\emph{arXiv preprint arXiv:1801.00690}}
  (\bibinfo{year}{2018}).
\newblock


\bibitem[\protect\citeauthoryear{Tjanaka, Fontaine, Togelius, and
  Nikolaidis}{Tjanaka et~al\mbox{.}}{2022}]%
        {tjanaka2022approximating}
\bibfield{author}{\bibinfo{person}{Bryon Tjanaka}, \bibinfo{person}{Matthew~C
  Fontaine}, \bibinfo{person}{Julian Togelius}, {and} \bibinfo{person}{Stefanos
  Nikolaidis}.} \bibinfo{year}{2022}\natexlab{}.
\newblock \showarticletitle{Approximating Gradients for Differentiable Quality
  Diversity in Reinforcement Learning}.
\newblock \bibinfo{journal}{\emph{arXiv preprint arXiv:2202.03666}}
  (\bibinfo{year}{2022}).
\newblock


\bibitem[\protect\citeauthoryear{Verhellen and Van~den Abeele}{Verhellen and
  Van~den Abeele}{2020}]%
        {verhellen2020illuminating}
\bibfield{author}{\bibinfo{person}{Jonas Verhellen} {and}
  \bibinfo{person}{Jeriek Van~den Abeele}.} \bibinfo{year}{2020}\natexlab{}.
\newblock \showarticletitle{Illuminating elite patches of chemical space}.
\newblock \bibinfo{journal}{\emph{Chemical science}} \bibinfo{volume}{11},
  \bibinfo{number}{42} (\bibinfo{year}{2020}), \bibinfo{pages}{11485--11491}.
\newblock


\bibitem[\protect\citeauthoryear{Vinyals, Babuschkin, Czarnecki, Mathieu,
  Dudzik, Chung, Choi, Powell, Ewalds, Georgiev, et~al\mbox{.}}{Vinyals
  et~al\mbox{.}}{2019}]%
        {vinyals2019grandmaster}
\bibfield{author}{\bibinfo{person}{Oriol Vinyals}, \bibinfo{person}{Igor
  Babuschkin}, \bibinfo{person}{Wojciech~M Czarnecki},
  \bibinfo{person}{Micha{\"e}l Mathieu}, \bibinfo{person}{Andrew Dudzik},
  \bibinfo{person}{Junyoung Chung}, \bibinfo{person}{David~H Choi},
  \bibinfo{person}{Richard Powell}, \bibinfo{person}{Timo Ewalds},
  \bibinfo{person}{Petko Georgiev}, {et~al\mbox{.}}}
  \bibinfo{year}{2019}\natexlab{}.
\newblock \showarticletitle{Grandmaster level in StarCraft II using multi-agent
  reinforcement learning}.
\newblock \bibinfo{journal}{\emph{Nature}} \bibinfo{volume}{575},
  \bibinfo{number}{7782} (\bibinfo{year}{2019}), \bibinfo{pages}{350--354}.
\newblock


\bibitem[\protect\citeauthoryear{Wang, Lehman, Clune, and Stanley}{Wang
  et~al\mbox{.}}{2019}]%
        {wang2019poet}
\bibfield{author}{\bibinfo{person}{Rui Wang}, \bibinfo{person}{Joel Lehman},
  \bibinfo{person}{Jeff Clune}, {and} \bibinfo{person}{Kenneth~O Stanley}.}
  \bibinfo{year}{2019}\natexlab{}.
\newblock \showarticletitle{Poet: open-ended coevolution of environments and
  their optimized solutions}. In \bibinfo{booktitle}{\emph{Proceedings of the
  Genetic and Evolutionary Computation Conference}}. \bibinfo{pages}{142--151}.
\newblock


\bibitem[\protect\citeauthoryear{Wang, Lehman, Rawal, Zhi, Li, Clune, and
  Stanley}{Wang et~al\mbox{.}}{2020}]%
        {wang2020enhanced}
\bibfield{author}{\bibinfo{person}{Rui Wang}, \bibinfo{person}{Joel Lehman},
  \bibinfo{person}{Aditya Rawal}, \bibinfo{person}{Jiale Zhi},
  \bibinfo{person}{Yulun Li}, \bibinfo{person}{Jeffrey Clune}, {and}
  \bibinfo{person}{Kenneth Stanley}.} \bibinfo{year}{2020}\natexlab{}.
\newblock \showarticletitle{Enhanced poet: Open-ended reinforcement learning
  through unbounded invention of learning challenges and their solutions}. In
  \bibinfo{booktitle}{\emph{International Conference on Machine Learning}}.
  PMLR, \bibinfo{pages}{9940--9951}.
\newblock


\end{thebibliography}
